\documentclass{rifpaper_arxiv}
\addauthor{Nicholas Baskerville}{n.p.baskerville@bristol.ac.uk}{School of Mathematics, University of Bristol}

\usepackage[english]{babel}
\usepackage[utf8]{inputenc}
\usepackage{graphicx}
\usepackage[colorinlistoftodos]{todonotes}
\usepackage{mathtools}
\usepackage{siunitx}
\usepackage{float}
\usepackage{textcomp}
\usepackage{tcolorbox}

\usepackage{environ}
\usepackage{listings}
\usepackage[makeroom]{cancel}
\usepackage{censor}
\usepackage[ruled,vlined]{algorithm2e}
\usepackage{algorithmic}
\usepackage{subcaption}
\usepackage{csquotes}


\usepackage{todonotes}

\usepackage{tabularx}
  \newcolumntype{L}{>{\raggedright\arraybackslash}X}

\renewcommand{\vec}[1]{\underline{#1}}
\usepackage{booktabs}
\title{A novel sampler for Gauss-Hermite determinantal point processes with application to Monte Carlo integration}

\date{\today}

\renewcommand{\vec}[1]{\bm{#1}}
\newcommand{\n}{\bm{n}}
\newcommand{\m}{\bm{m}}
\newcommand{\x}{\bm{x}}
\newcommand{\R}{\mathbb{R}}
\newcommand{\N}{\mathbb{N}}
\newcommand{\fb}{\mathfrak{b}}
\newcommand{\Tr}{\text{Tr}}

\begin{document}

\maketitle
\begin{abstract}
    Determinantal points processes are a promising but relatively under-developed tool in machine learning and statistical modelling, being the canonical statistical example of distributions with repulsion.
    While their mathematical formulation is elegant and appealing, their practical use, such as simply sampling from them, is far from straightforward.
    Recent work has shown how a particular type of determinantal point process defined on the compact multidimensional space $[-1, 1]^d$ can be practically sampled and further shown how such samples can be used to improve Monte Carlo integration.
    This work extends those results to a new determinantal point process on $\R^d$ by constructing a novel sampling scheme.
    Samples from this new process are shown to be useful in Monte Carlo integration against Gaussian measure, which is particularly relevant in machine learning applications.
\end{abstract}

\section{Introduction}
Determinantal point processes (DPP) define distributions over sets of points in a metric space, e.g. Euclidean $\R^d$, which are characterised by repulsion between points.
An archetypal example from random matrix theory are the eigenvalues of the Gaussian unitary ensemble (GUE) \cite{mehta2004random}.
Their joint p.d.f. can be shown to be \cite{livan2018introduction,mehta2004random} \begin{align}
    p(\lambda_1, \ldots, \lambda_N) \propto \prod_{i < j} |\lambda_i - \lambda_j|^2 \prod_{i=1}^N \frac{e^{-\frac{\lambda_i^2}{2}}}{\sqrt{2\pi}}.
\end{align}
One sees immediately the competing effects of the repulsive terms $|\lambda_i-\lambda_j|^2$ which encourage configurations of eigenvalues with large pairwise separations and the confining Gaussian potentials $e^{-\frac{\lambda_i^2}{2}}$ which encourage eigenvalues to stray not too far from the origin.
The eigenvalues of a GUE and other random matrix ensembles can indeed be seen quite precisely to be distributed as certain determinantal point processes \cite{mehta2004random}

\medskip
In general a DPP on locally compact Polish space\footnote{Recall that a Polish space is separable and completely metrizable. Separability (i.e. the existence of some dense countable subset) is crucial for a well-behaved product measure in this context. Being completely metrizable makes speaking of convergence much simpler. Local compactness is a natural condition when one is speaking of repulsion between points.} $\mathbb{X}$ is defined by a measure $\mu$ and a measurable \emph{kernel} $K$ such that, for any $n\in\N$ \begin{align}
    p(x_1, \ldots x_n)dx_1\ldots dx_n = \frac{1}{N!}\det\left(K(x_i, x_j)\right)_{i,j=1}^n \prod_{i=1}^n \mu(dx_i).
\end{align}
The repulsive nature of the DPP is encoded in the determinantal structure.
Intuitively, the determinant measures the volume occupied by the points $x_i$ embedded in some space induced by the kernel and so samples are biased towards configurations with larger volumes, meaning greater separation between points.
More precisely, consider a \emph{projection kernel} \cite{hough2006determinantal} defined as
\begin{align}
    K_N(x, x') = \frac{1}{N} \sum_{i=1}^N\phi_i(x)\phi_i(x')
\end{align}
where $\phi_i$ are orthonormal functions on $\mathbb{X}$ with respect to the chosen measure $\mu$, i.e.
\begin{align}\label{eq:projection_def}
    \int_{\mathbb{X}} d\mu(x) \phi_i(x)\phi_j(x) = \delta_{ij}.
\end{align}
With this construction \begin{align}
    \det\left(K(x_i, x_j)\right)_{i,j=1}^n = \left(\det\vec{\Phi}(x_{1:N})\right)^2
\end{align}
where the matrix $\vec{\Phi}(x_{1:N})_{ij} = \phi_j(x_i)$ and so we see that the DPP density favours configurations with large separation of the $x_i$ in the embedding space induced by the $\phi_j$.
Projection DPPs can be viewed as a truncation of DPPs.
Indeed by Mercer's theorem there exists some eigenfunction expansion
\begin{align}
    K(x, y) = \sum_{i=1}^{\infty} \lambda_i \phi_i(x)\phi_i(y).
 \end{align}
Truncating this expansion and rescaling appropriately to absorb the eigenvalues produces the projection DPP expression (\ref{eq:projection_def}).
A DPP so defined is referred to as $\text{DPP}(\mu, K_N)$.
Samples from $\text{DPP}(\mu, K_N)$ have the convenient property that their cardinality is almost surely $N$ \cite{hough2006determinantal,NEURIPS2019_1d54c76f}

\medskip
DPPs have several applications in machine learning and statistics \cite{kulesza2012determinantal}, being the natural class of distributions for any point process with repulsion.
Recently \cite{NEURIPS2019_1d54c76f} proposed using certain DPPs to construct Monte Carlo (MC) integration estimators that can outperform na\"{i}ve MC integration.
In particular, the authors discuss the Bardenet-Hardy (BH) estimator \cite{bardenet2020monte} and the Ermakov-Zolotukhin (EZ) estimator \cite{ermakov1960polynomial}.
Both estimators rely on on the repulsive nature of projection DPPs to provide super-na\"{i}ve MC convergence.
Indeed, \cite{NEURIPS2019_1d54c76f} establish a central limit theorem for BH with decay of speed $\sqrt{N^{1 + 1/d}}$; similar results for EZ are not yet available.

\medskip
\cite{NEURIPS2019_1d54c76f} focusses specifically on the case $\mathbb{X}=[-1, 1]^d$ with measure \begin{align}
    \mu(\x) = \prod_{i=1}^d (1-x)^{a_i}(1+x)^{b_i}dx_i
\end{align}
in which case the appropriate orthogonal functions $\phi_i$ are multivariate Jacobi polynomials.

\medskip
In this paper we extend approach of \cite{NEURIPS2019_1d54c76f} to Gaussian measure on $\R^d$.
In machine learning applications, accurate MC estimation of integration against Gaussian measure is arguably the most common case, and certainly more common than Jacobi measure.
Our techniques also open the way for applications of Gaussian DPPs on $\R^d$ in statistical models.
The central contribution we present is a novel rejection sampler for Gauss-Hermite projection DPPs based on a factorisation trick and a random matrix theory connection.
Alongside the development of this sampler, we also present experimental results using it in BH and EZ estimation, as well as timing and efficiency results for the sampler.

\section{Multivariate Hermite polynomials and a factorised DPP sampler}

\paragraph{Special case.}  In the one dimensional case, the procedure in this section is not required.
Rather one can sample a $N\times N$ GUE matrix $X$ and compute its eigenvalues $\lambda_1,\ldots, \lambda_N$.
The eigenvalues are then distributed as $\text{DPP}(\mu, K_N)$.

\subsection{Factorised sampler in special cases}
Let $\{H_j\}_{j=0}^{\infty}$ denote the Physicist's Hermite polynomials defined by the orthogonality condition on $\R$ \begin{align}
    \int_{\R} dx ~ e^{-x^2} H_j(x)H_k(x) = \sqrt{\pi}2^j j! \delta_{jk}.
\end{align}
We define from these the following orthonormal polynomials
\begin{align}
    \psi_j = \frac{H_j(x/\sqrt{2})}{\sqrt{\sqrt{2\pi}2^j j!}}
\end{align}
which obey \begin{align}
    \int_{\R} dx ~ e^{-\frac{x^2}{2}} \psi_j(x)\psi_k(x) = \delta_{jk}.
\end{align}
From these $\psi_j$ it is simple to define multivariate polynomials orthonormal with respect to Gaussian measure on $\R^d$
\begin{align}
    \phi_{\vec{i}}(\vec{x}) = \prod_{\ell = 1}^d \psi_{i_{\ell}}(x_{\ell})
\end{align}
where $\vec{i}$ is a multi-index of length $d$ and indeed 
\begin{align}
    \int_{\R^d} d\x ~ e^{-\frac{\x^2}{2}} \phi_{\vec{j}}(\x)\phi_{\vec{k}}(\x) = \prod_{\ell=1}^d \int_{\R} dx_{\ell} ~ e^{-\frac{x_{\ell}^2}{2}} \psi_{j_{\ell}}(x_{\ell})\psi_{k_{\ell}}(x_{\ell}) = \prod_{\ell=1}^d\delta_{j_{\ell}k_{\ell}} = \delta_{\vec{j}\vec{k}}.
\end{align}
Following \cite{NEURIPS2019_1d54c76f} we can place a natural ordering $\fb:\N^d\rightarrow\N$ on the $\{\phi_{\vec{i}}\}$ sorting first by the maximum degree of $\vec{i}$ and within a given maximum degree sorting lexicographically.
Again following \cite{NEURIPS2019_1d54c76f,hough2006determinantal} we construct a kernel from a finite dimensional subspace of $\text{span}(\{\phi_{\vec{i}}\}_{\fb(\vec{i})=0}^{\infty})$:
\begin{align}
    K_N(\vec{x}, \vec{y}) = \sum_{\fb(\vec{i})=0}^{N-1} \phi_{\vec{i}}(\vec{x})\phi_{\vec{i}}(\vec{y}).
\end{align}
We can now define a projection DPP with respect to un-normalised standard Gaussian base measure on $\R^d$, denoted by $\mu$, i.e. $d\mu(\x) = e^{-\frac{\x^2}{2}}d\x$.
\begin{align}
    \{\x_1, \ldots, \x_N\}\sim\text{DPP}(\mu, K_N) ~ \iff ~ p(\x_1, \ldots, \x_N)d\x = \frac{1}{N!}\det\left(K_N(\x_p, \x_q)\right)_{p,q=1}^N d\mu(\x).
\end{align}
To sample from $\text{DPP}(\mu, K_N)$ we can follow a procedure similar to that used in \cite{NEURIPS2019_1d54c76f} for DPPs with respect to Jacobi base measure.
\cite{NEURIPS2019_1d54c76f} gives the follows factorisation
\begin{align}\label{eq:dpp_factorisation}
    p(\x_1, \ldots, \x_N)d\x_1\ldots d\x_N = \frac{K_N(\x_1, \x_1)}{N} \omega(\x_1) d\x_1 \prod_{n=2}^N \omega(\x_n)d\x_n \frac{K_N(\x_n, \x_n) - \vec{k}_{n-1}(\x_n)^T\vec{K}_{n-1}^{-1}\vec{k}_{n-1}(\x_n)}{N - (n-1)},
\end{align}
where $\omega(\x) = e^{-\frac{\x^2}{2}}$, $\vec{k}_{n-1}(\x) = (K_N(\x_1, \x), \ldots, K_N(\x_{n-1}, \x))^T$ and $(\vec{K}_{n})_{ij} = K_N(\vec{x}_i, \vec{x}_j)$ for $1 \leq i,j \leq n$.
This factorisation follows from writing the $N\times N$ kernel matrix as a $( (N-1) + 1) )\times ( (N-1) + 1) )$ block matrix, separating out $\vec{K}_{N-1}$ from $K_N(\vec{x}_N, \vec{x}_N)$, using the Schur formula for block matrix determinants, and finally iterating this procedure.
Thus for a given $N$, points are sampled sequentially using a chain rule scheme i.e. $\x_j$ is sampled from $p(\x_j \mid \x_1,\ldots, \x_{j-1})$ by exploiting the above factorisation.
The artificial induced order can then been thrown away to give a sample from the DPP.
The problem is thus reduced too sampling from the conditional densities given in the product in (\ref{eq:dpp_factorisation}).
A rejection sampler with good rejection bound can be used for the conditionals $p(\x_j \mid \x_1,\ldots, \x_{j-1})$ using as proposal distribution 
\begin{align}
    f_N(\x)d\x = \frac{1}{N} d\mu(\x)\sum_{\fb(\vec{i})=0}^{N-1} \phi_{\vec{i}}(\x)^2 = \frac{1}{N} d\mu(\x) K_N(\x, \x).
\end{align}
Indeed, one has \begin{align}
    \frac{K_N(\x_n, \x_n) - \vec{K}_{n-1}(\x_n)^T\vec{K}_{n-1}^{-1}\vec{K}_{n-1}(\x_n)}{N - (n-1)} \left(f(\x)\right)^{-1} \leq \frac{N}{N - (n-1)}
\end{align}
which follows from the positive definiteness of kernel matrices and holds for any choice of kernel and any base measure $\mu$.
\medskip
At this point \cite{NEURIPS2019_1d54c76f} sample from $f_N(\x)d\x$ by viewing it as a mixture and constructing efficient rejection samplers for the mixture components with distributions $d\mu(\x)\phi_{\vec{i}}(\x)^2$.
This approach cannot be adapted to the Gauss-Hermite case considered here, as it relies on good uniform bounds which are not available for Hermite polynomials\footnote{See \cite{krasikov2004new} for the best bounds on Hermite polynomials. Roughly speaking, the problem is that $\psi_n(x)^2$ has very sharp peak values which grow with $n$. See also Appendix \ref{app:spherical} for an alternative Gaussian orthonormal function basis that also suffers from bad bounds.} but are for Jacobi polynomials.
We propose an alternative sampler for $f_N(\x)d\x$ that side-steps the need for uniform bounds on Hermite polynomials and instead exploits a connection to random matrix theory.
Note that our approach could be applied, with appropriate modifications, to other DPPs with different base measures, including the Jacobi DPP studied in \cite{bardenet2020monte}.
Initially we specialise to the case $N = n^d$ where $n$ is a positive integer. %, and in practice we will assume that $n$ is sufficiently large, say $n \geq 5$.
With this assumption one can make the following factorisation
\begin{align}
    f_N(\x) = \prod_{\ell=1}^d \left(\frac{1}{n} e^{-\frac{x_{\ell}^2}{2}} \sum_{i_{\ell}=0}^{n} \psi_{i_{\ell}}(x_{\ell})^2\right)
\end{align}
and so sampling from $f_N(\x)d\x$ can be achieved by sampling the entries of $\x$ \emph{independently} from \begin{align}
    \rho_{n}(x_{\ell})dx_{\ell} = \frac{1}{n} e^{-\frac{x_{\ell}^2}{2}} \sum_{i_{\ell}=0}^{n} \psi_{i_{\ell}}(x_{\ell})^2 ~ dx_{\ell}.
\end{align}
To sample from the univariate distribution $\rho_{n}(x)dx$ we draw on a result from random matrix theory, namely that $\rho_{n}$ is the spectral density of an $n\times n$ Gaussian unitary ensemble (GUE) matrix \cite{livan2018introduction}.
An Hermitian $n\times n$ random matrix $X$ is GUE if it has density proportional to $e^{-\frac{1}{2}\Tr X^2}$.
The eigenvalues and eigenvectors of a GUE matrix are independent so let us denote the density of its unordered eigenvalues by $p(\lambda_1, \ldots,\lambda_n)$.
The \emph{spectral density} is then simply the single eigenvalue marginal \begin{align}
    \int dx_2\ldots dx_{n} ~ p(x, x_2, \ldots, x_n).
\end{align}
This suggests the following sampling scheme:
\begin{enumerate}
    \item Sample $X$ as a $n\times n$ GUE matrix.
    \item Compute the eigenvalues $\lambda_1, \ldots, \lambda_{n}$ of $X$.
    \item Select $x$ uniformly at random from $\{\lambda_1, \ldots, \lambda_{n}\}$.
    \item $x$ has distribution $\rho_{n}(x)dx$.
\end{enumerate}

While elegant, this scheme is quite inefficient. Sampling an $n\times n$ GUE matrix requires sampling $n^2$ independent standard univariate Gaussians. 
Computing the spectrum of $X$ is an $\mathcal{O}(n^3)$ operation. So overall each sample drawn from $\rho_{n}(x)dx$ by this method has time complexity $\mathcal{O}(n^3) + n^2$.
The random matix theory interpretation of $\rho_{n}$ does however provide an alternative sampling approach.
The following result is a special case of Wigner's theorem \cite{livan2018introduction}:
\begin{align}\label{eq:wigner_thm}
    \lim_{n\rightarrow\infty} \sqrt{2n}\rho(\sqrt{2n}x) = \frac{1}{\pi} \sqrt{2 - x^2} ~\mathbb{I}\{x^2 \leq 2\} \equiv \rho_{SC}(x)
\end{align}
where the density on the right hand side is Wigner's semi-circle.
The error terms in (\ref{eq:wigner_thm}) are actually known to be $\mathcal{O}(n^{-1})$, which can be seen easily from the Christoffel-Darboux expansions in Section 10.4 of \cite{livan2018introduction}.
We can thus say \begin{align}\label{eq:semicircle_approx}
    \rho_{n}(x) = \frac{1}{2 n\pi} \sqrt{4n - x^2} ~ \mathbb{I}\{ x^2 \leq 4n\} + \mathcal{O}(n^{-3/2})
\end{align}
and the error terms are non-vacuous in the bulk of the support excluding regions of width $\mathcal{O}(n^{-3/2})$ at the edges of the support.
This approximation suggests a rejection sampling approach. Consider the following class of mixture distributions with densities
\begin{align}
    h_{p, \nu}(x) = pg_{\nu}(x) + (1-p)\rho_{SC}(x)
\end{align}
where $g_{\nu}$ is Student's $t$ density with $\nu$ degrees of freedom. We will use $h_{p, \nu}$ as the proposal distribution for a rejection sampler of $\rho_n(x)dx$.
The reasoning behind this choice is the following:
\begin{itemize}
    \item (\ref{eq:semicircle_approx}) suggests using a semi-circle as proposal, i.e. propose $x = (\epsilon + \sqrt{2n})y$ where $y\sim \rho_{SC}(x)dx$ and $\epsilon>0$ is small and included to inflate the semi-circle slightly so that it encompasses $\rho_n$.
    \item This approach fails however, as $\rho_n$ is in fact supported on all of $\R$.
    \item To ensure that any rejection bound holds in the tails, we seek to modify the semi-circle to endow it with heavy tails.
    For large $n$, the largest eigenvalue of a $n\times n$ GUE follows the Tracy-Widom tail distribution which has a right tail like $e^{-cx^{3/2}}$ \cite{anderson2010introduction}.
    This heuristic suggests that a proposal with Gaussian tails may not have sufficient tail weight.
    \item Hence the mixture of a semi-circle and a Student's $t$. It may be possible to use a lighter-tailed distribution, however we find this is a stable and reliable choice, ensuring that the rejection sampler bound is satisfied everywhere.
\end{itemize}

\paragraph{Sampling from the semi-circle.}To use $h_{p, \nu}$ as proposal density, we must be able to sample from it.
Sampling from Student's $t$ is standard, so we need only address sampling from the semi-circle.
The semi-circle is compactly supported and has a simple unimodal shape, so it is easy to hand-engineer an appropriate rejection sampler.
We choose a proposal density
\begin{align}
    r(x) = \frac{1}{2}(\mathcal{N}(x; -1, (4/5)^2) + \mathcal{N}(x; 1, (4/5)^2))
\end{align}
and then numerically find the bound \begin{align}
    \frac{\rho_{SC}(x)}{r(x)} \leq 1.42.
\end{align}

\medskip
We of course require a bound $\rho_{n}/h_{p, \nu} \leq M_{n} ~ \forall~x$ so that rejection sampling can be used.
Such a bound does not appear analytically forthcoming, however since we are a considering only simple univariate densities, it is quite practical to hand-craft a bound empirically.
Indeed, it is a simple matter numerically to compute near-optimal bounds for any $n, p$ and $\nu$.
The acceptance rate can then be estimated and finally the best values $p_{n}, \nu_{n}$ along with near-optimal bounds $M_{n}$ can be found for each $n$ to give near optimal acceptance rates.

\begin{figure}
    \centering
    \includegraphics[width=0.4\linewidth]{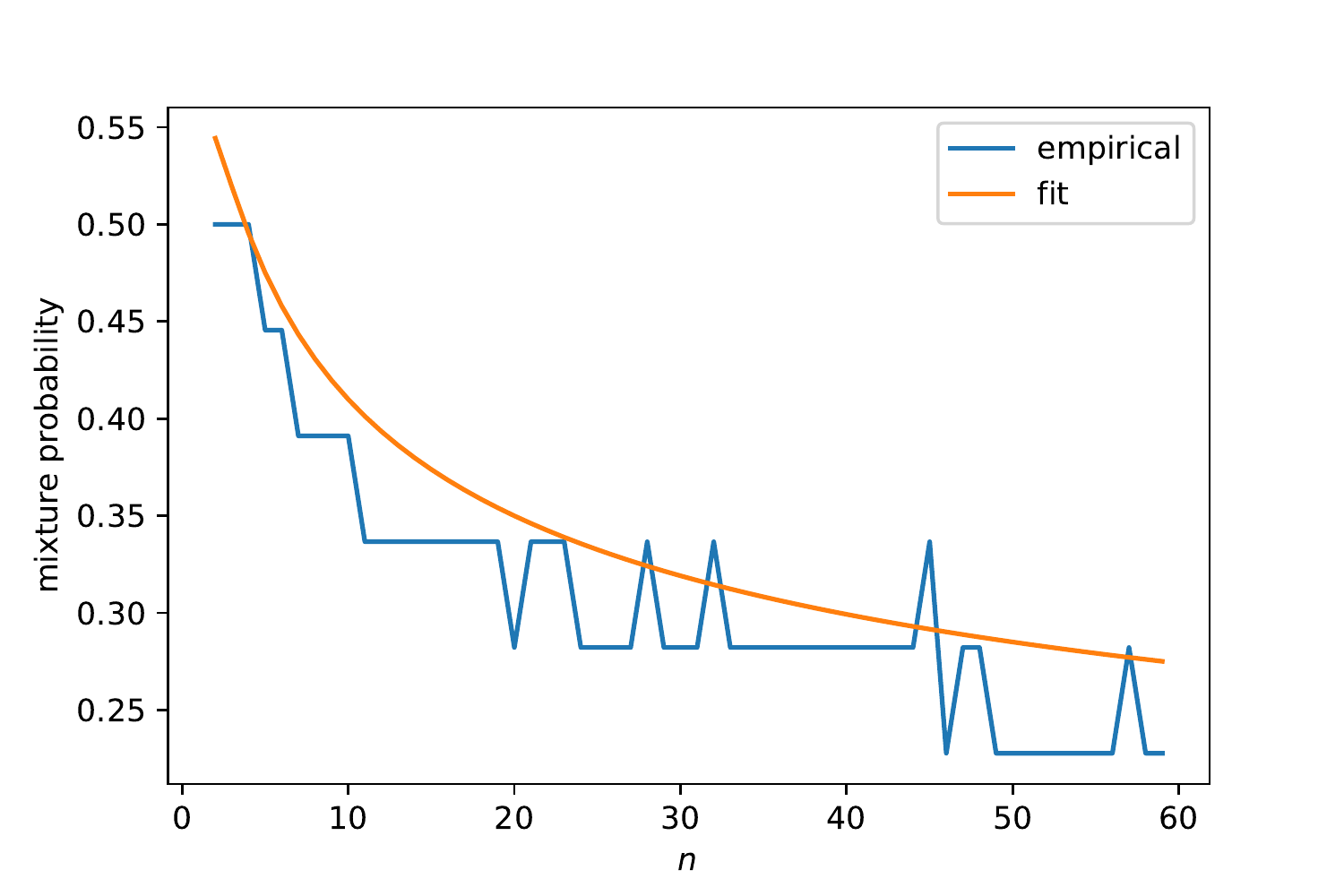}
    \caption{Empirical results for the optimal mixture probability with respect to the best attainable acceptance probability in the rejection sampler for $\rho_{n}$.}
    \label{fig:optimal_p}
\end{figure}

\medskip
Following the above empirical procedure, we conclude the degrees of freedom $\nu$ is not particularly important to the efficiency of the sampler, so we set it to 10.
The mixture probability $p$ is important, however, and we fit a simple parametric curve to it:
\begin{align}\label{eq:mixture_p}
    0.100 - \frac{0.486}{n} + \frac{0.647}{n^{1/2}} + \frac{0.272}{n^{1/4}}.
\end{align}
The empirical results and this simple fit are shown in Figure \ref{fig:optimal_p}.
Note that our parametric fit is monotonically decreasing, which is intuitively appropriate as the tails of $\rho_n$ lighten as $n$ increases.

\medskip
Now fixing the form of the mixture probability in (\ref{eq:mixture_p}), we numerically find approximations to the tightest bounds $M_{n}$. 
Practically, this is done by evaluating the ratio $\rho_{n}/h_{p, \nu}$ over $(-2\sqrt{n} - 10, 2\sqrt{n} + 10)$ on a finely-spaced grid and taking the maximum.
We then fit a parametric form to the resulting data, building in some extra tolerance to ensure that our parametric fit is certainly a uniform upper bound for $\rho_{n}/f_{p, \nu}$.
The results are shown in Figure \ref{fig:optimal_m} and the parametric fit is
\begin{align}\label{eq:reject_bound}
    0.492 + \frac{1.058}{n} - \frac{3.352}{n^{1/2}} + \frac{3.308}{n^{1/4}}.
\end{align}
\begin{figure}
    \centering
    \includegraphics[width=0.4\linewidth]{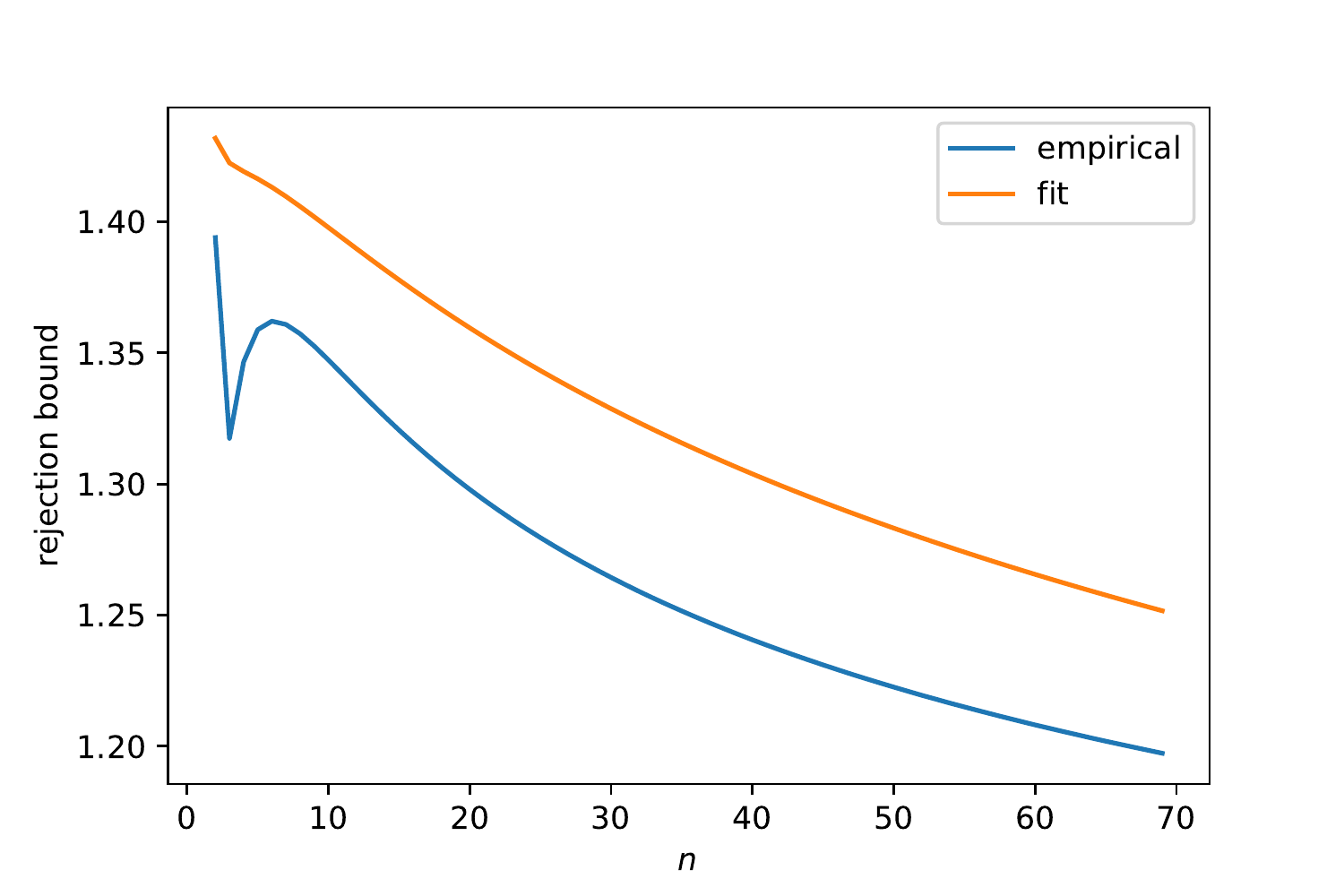}
    \caption{Empirical results for the tightest uniform rejection bound  on $\rho_{n}/f_{p, \nu}$.}
    \label{fig:optimal_m}
\end{figure}

\begin{figure}
    \centering
    \begin{subfigure}[b]{0.3\textwidth}
        \centering
        \includegraphics[width=\textwidth]{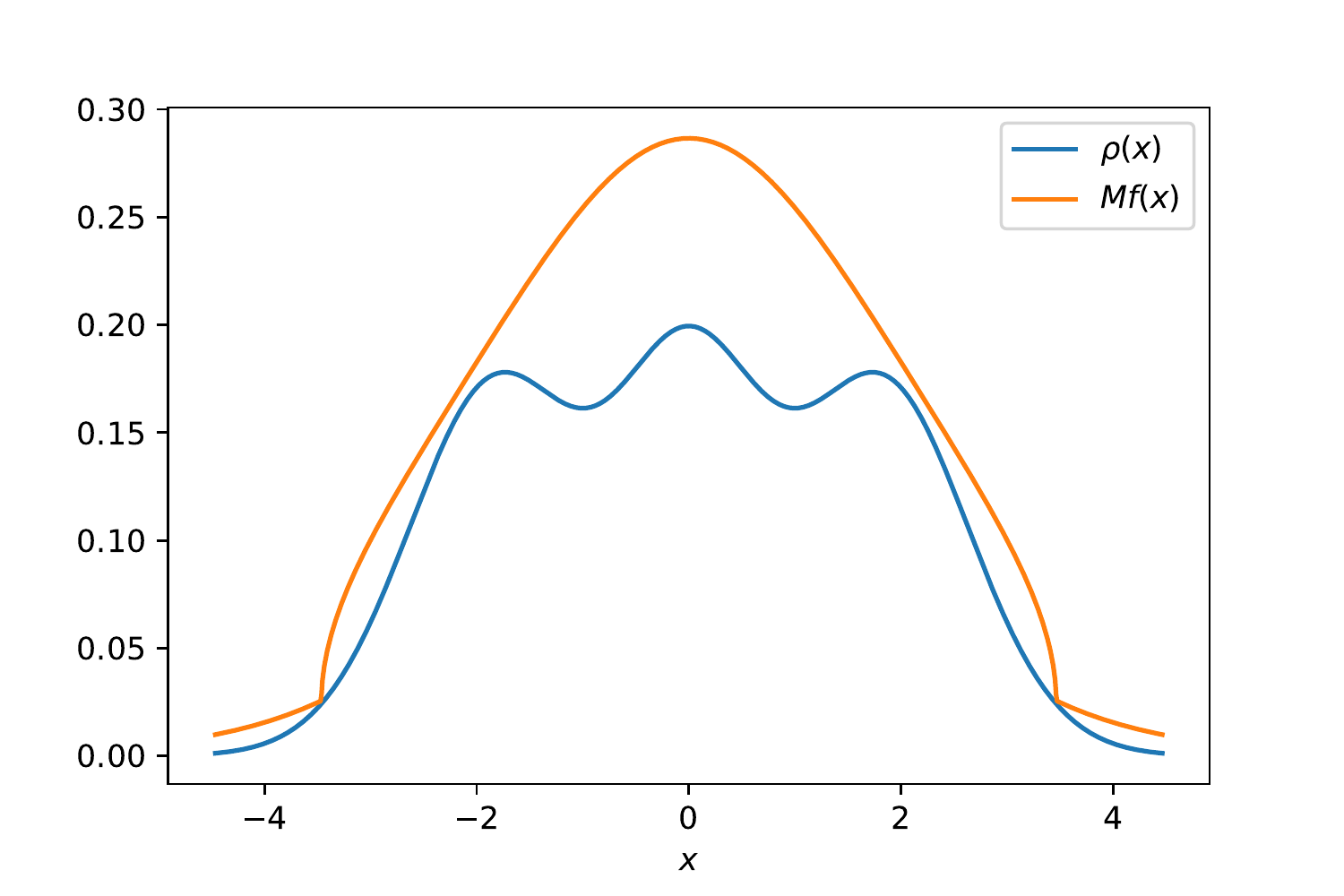}
        \caption{$n=3$}
    \end{subfigure}
    \begin{subfigure}[b]{0.3\textwidth}
        \centering
        \includegraphics[width=\textwidth]{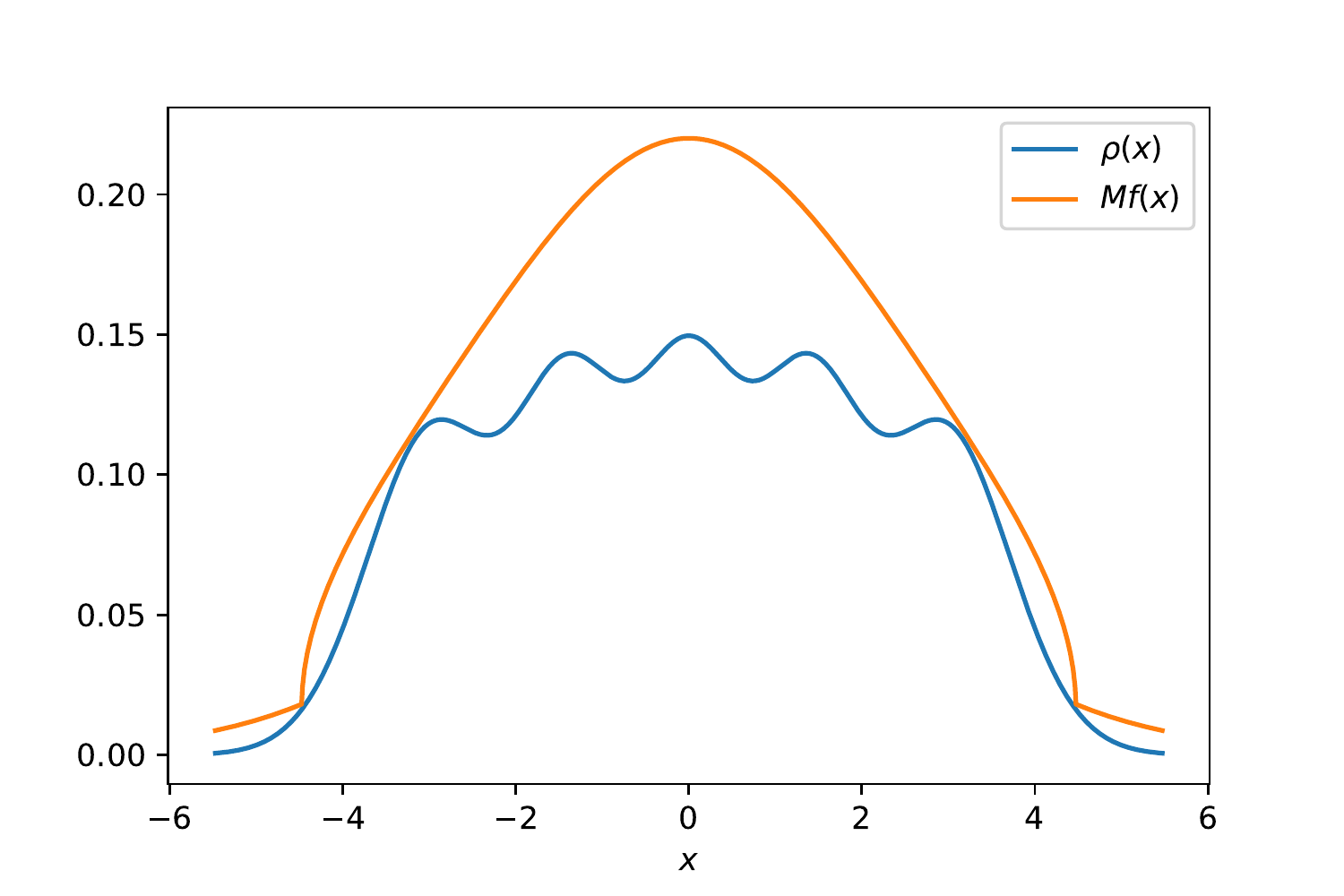}
        \caption{$n=5$}
    \end{subfigure}
    \begin{subfigure}[b]{0.3\textwidth}
        \centering
        \includegraphics[width=\textwidth]{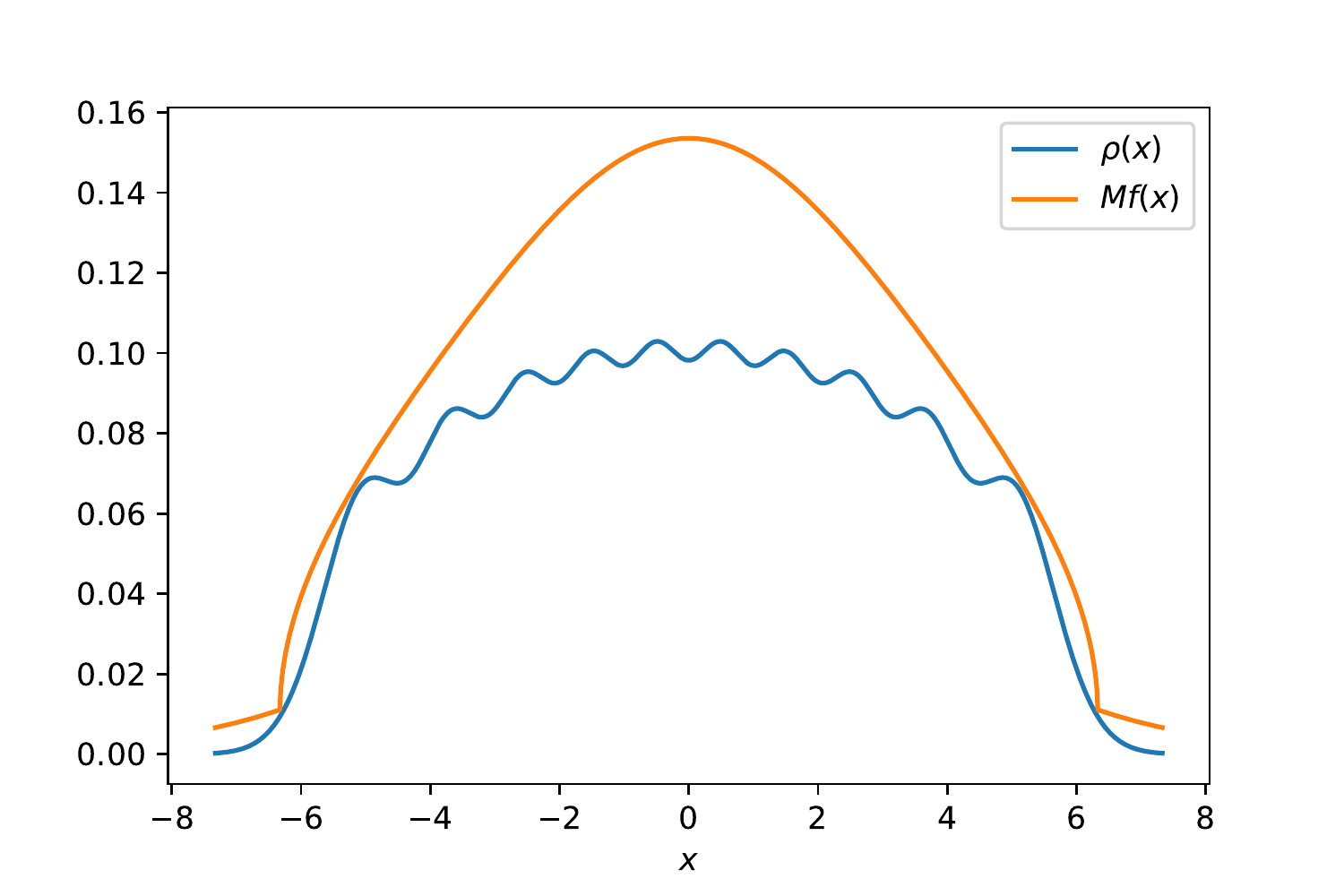}
        \caption{$n=10$}
    \end{subfigure}
    \caption{Comparison of the target density $\rho_n$ and the scaled proposal density $M_nf_{p_n, \nu}$ for various $n$ values.}
    \label{fig:rejection_sampler_bound}
\end{figure}

Figure \ref{fig:rejection_sampler_bound} shows the target density $\rho_n$ and the scaled proposal density $M_nh_{p_n, \nu}$.
We show some results of using our rejection sampler for $\rho_n$ for various $n$ in Figure \ref{fig:optimal_m}.

\begin{figure}
    \centering
    \begin{subfigure}[b]{0.3\textwidth}
        \centering
        \includegraphics[width=\textwidth]{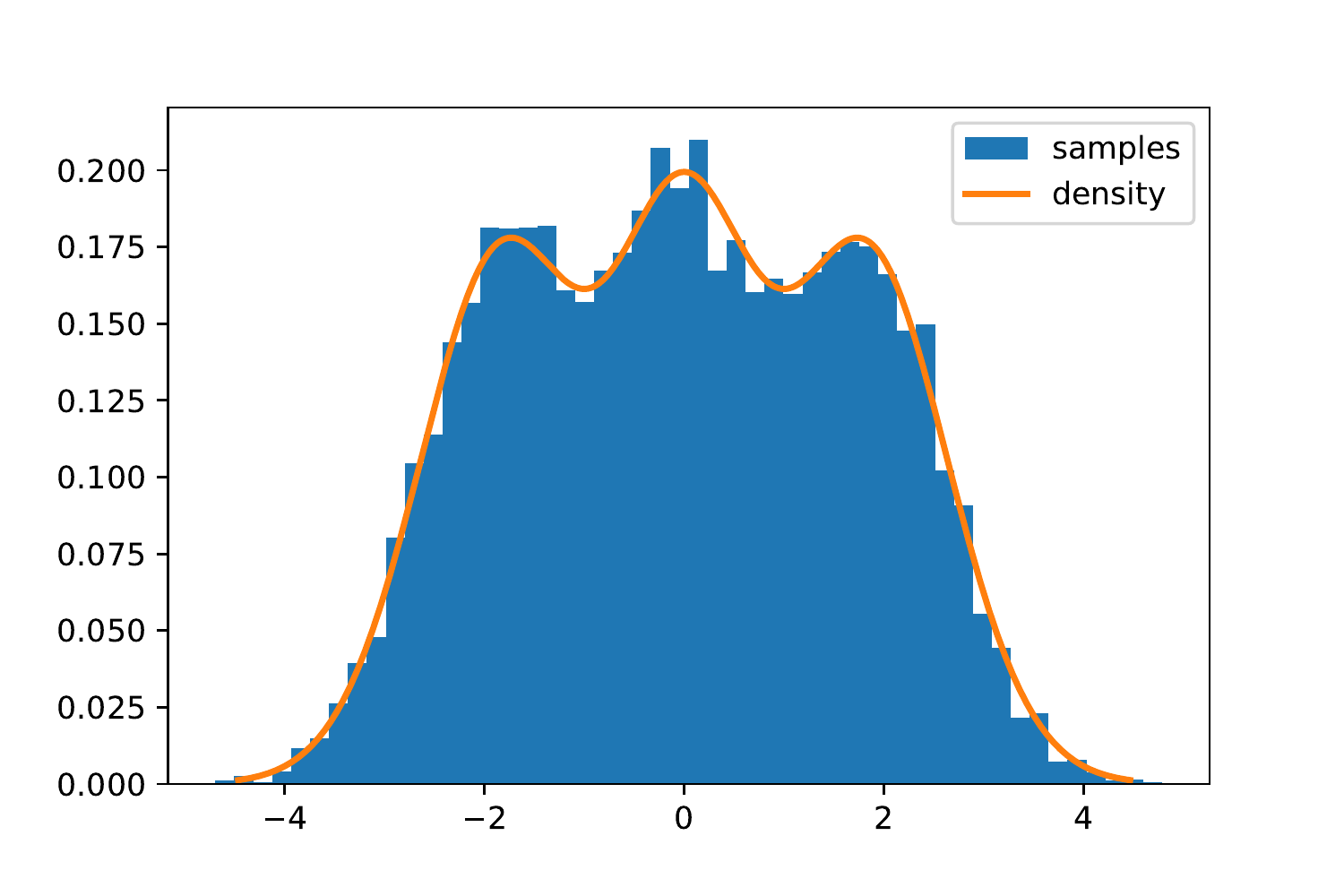}
        \caption{$n=3$}
    \end{subfigure}
    \begin{subfigure}[b]{0.3\textwidth}
        \centering
        \includegraphics[width=\textwidth]{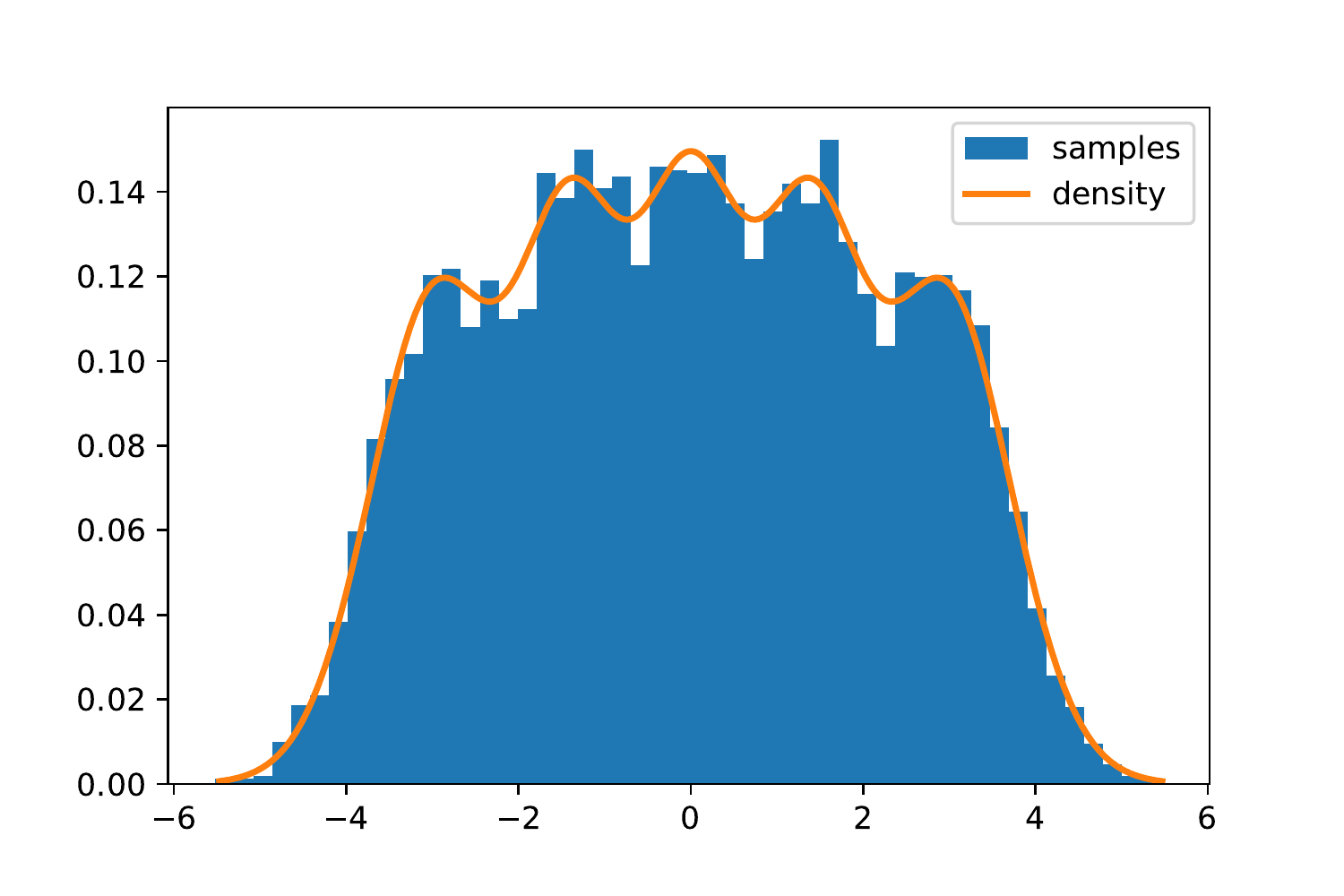}
        \caption{$n=5$}
    \end{subfigure}
    \begin{subfigure}[b]{0.3\textwidth}
        \centering
        \includegraphics[width=\textwidth]{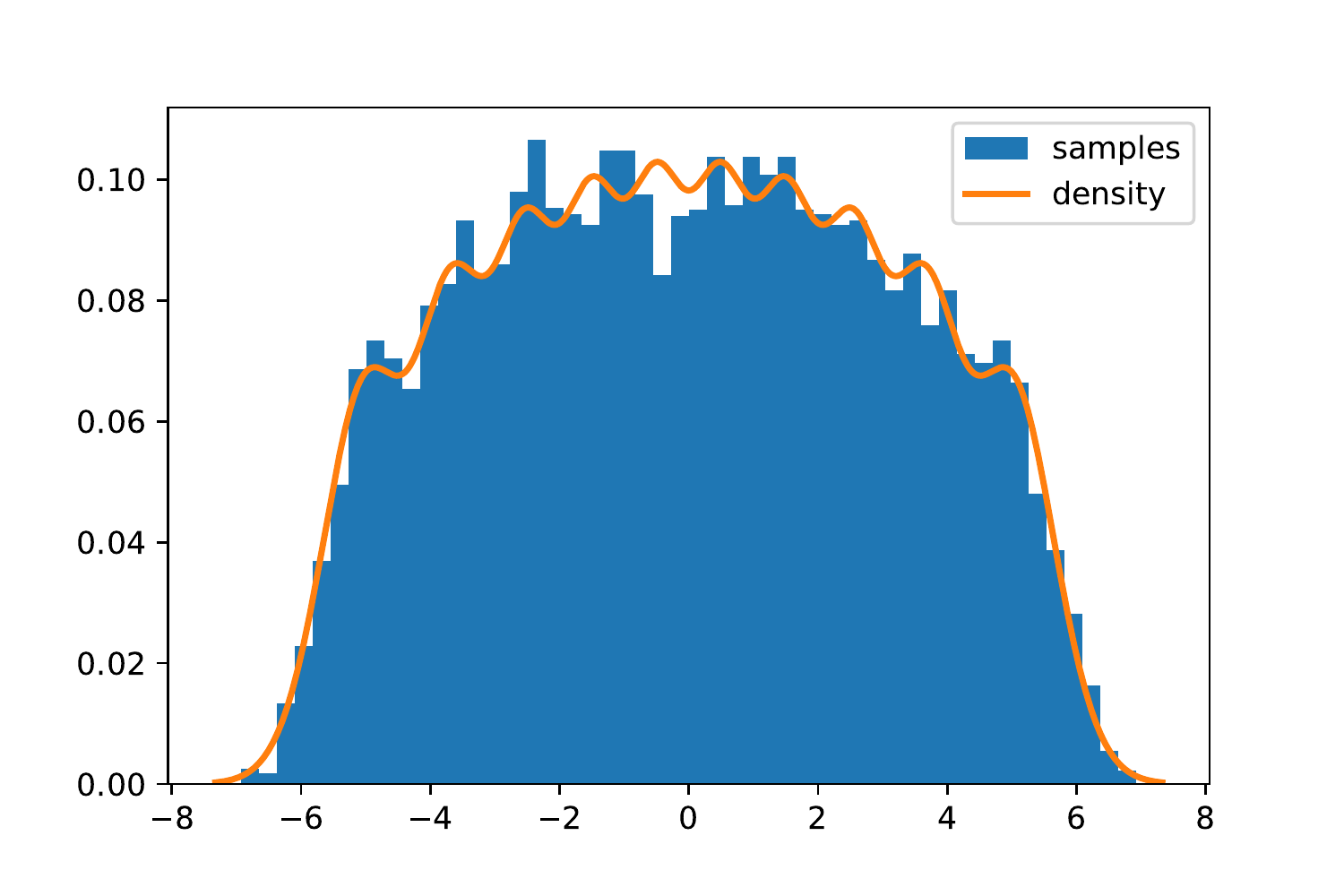}
        \caption{$n=10$}
    \end{subfigure}
    \caption{Results of using our rejection sampler for $\rho_n$ for a few different $n$ values. We used 10000 samples and 50 histogram bins.}
    \label{fig:optimal_m}
\end{figure}

\begin{figure}
    \centering
    \includegraphics[width=0.4\textwidth]{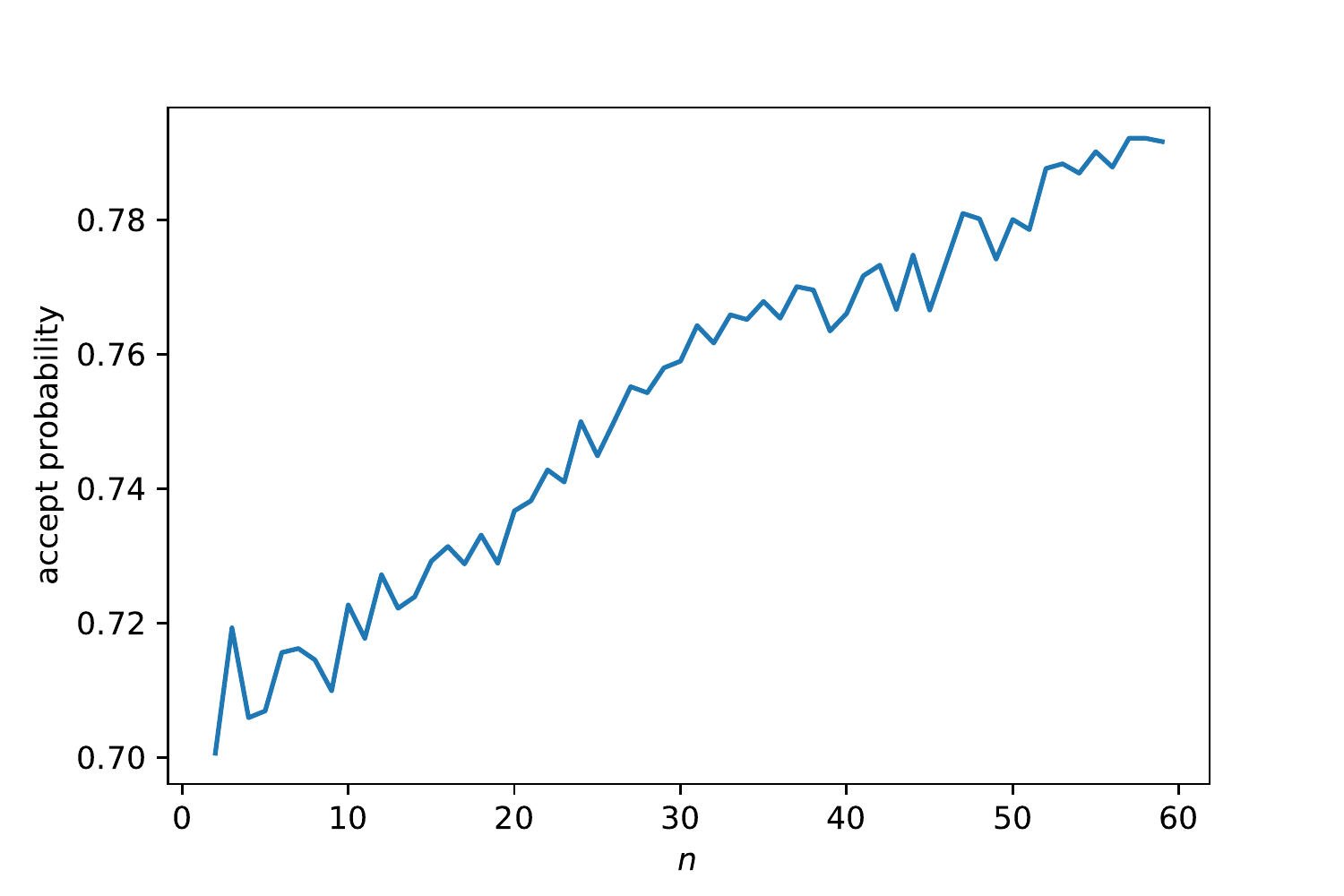}
    \caption{The empirically obtained acceptance probability for the rejection sampler of $\rho_n$ for various $n$ values. Calculated from 10000 trials.}
    \label{fig:accept_prob}
\end{figure}

Finally, fixing our parametric fits, we can estimate the actual obtained acceptance probability.
The results from 10000 trials for each $n$ value are shown in Figure \ref{fig:accept_prob}.
Note that the acceptance probability is generally increasing in $n$ and never less that $0.7$.

\medskip

\subsection{Extension to rejection sampler in the general case}\label{subsec:general_N}
In the previous subsection we restricted to an extremely special case $N=n^d$.
In this section we leverage the sampler we constructed in that case to construct one for general $N$.

\medskip
Given $N\in\N$, and dimension $d>1$, define $n = \lceil N^{1/d} \rceil$ and define also $m = n^d - N$.
Note that $0 \leq m \leq n^d - (n-1)^d - 1$. 
Define \begin{align}
    q_n(\x) = \frac{1}{n^d} d\mu(\x)\sum_{\fb(\vec{i})=0}^{n^d-1} \phi_{\vec{i}}(\x)^2
\end{align}
and then we can immediately write down the bound 
\begin{align}\label{eq:gen_N_reject}
 \frac{f_N(\x)}{q_n(\x)} = \frac{\frac{1}{N} d\mu(\x)\sum_{\fb(\vec{i})=0}^{N-1} \phi_{\vec{i}}(\x)^2}{\frac{1}{n^d} d\mu(\x)\sum_{\fb(\vec{i})=0}^{n^d-1} \phi_{\vec{i}}(\x)^2} \leq \frac{n^d}{n^d - m}.
\end{align}
Sampling from $q_n$ can be achieved using the factorised rejection sampler derived in the previous subsection, so we have all that is required for a rejection sampler for $f_N$ for general $N$.

\begin{figure}
    \centering
    \begin{subfigure}[b]{0.3\textwidth}
        \centering
        \includegraphics[width=\textwidth]{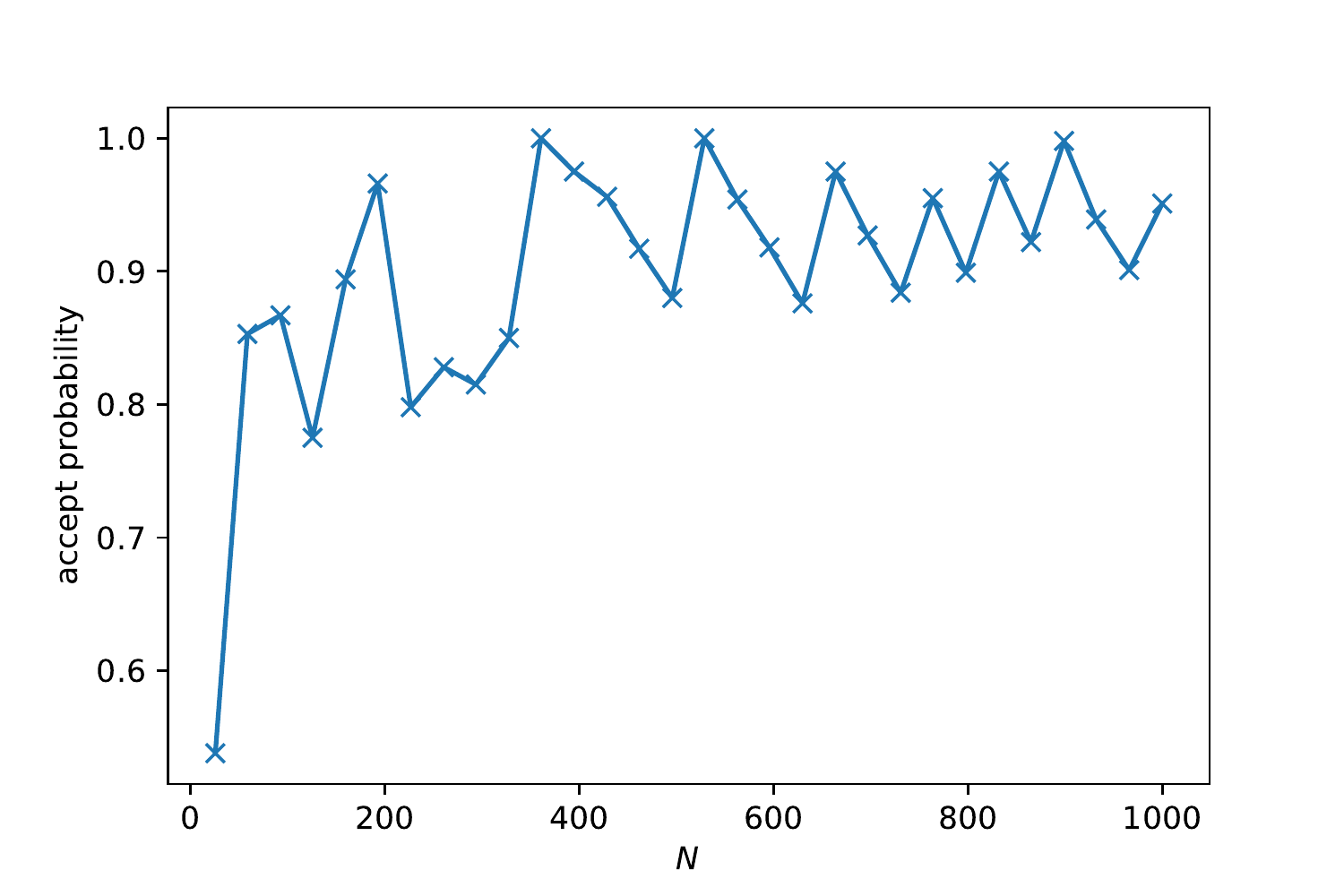}
        \caption{$d=2$}
    \end{subfigure}
    \begin{subfigure}[b]{0.3\textwidth}
        \centering
        \includegraphics[width=\textwidth]{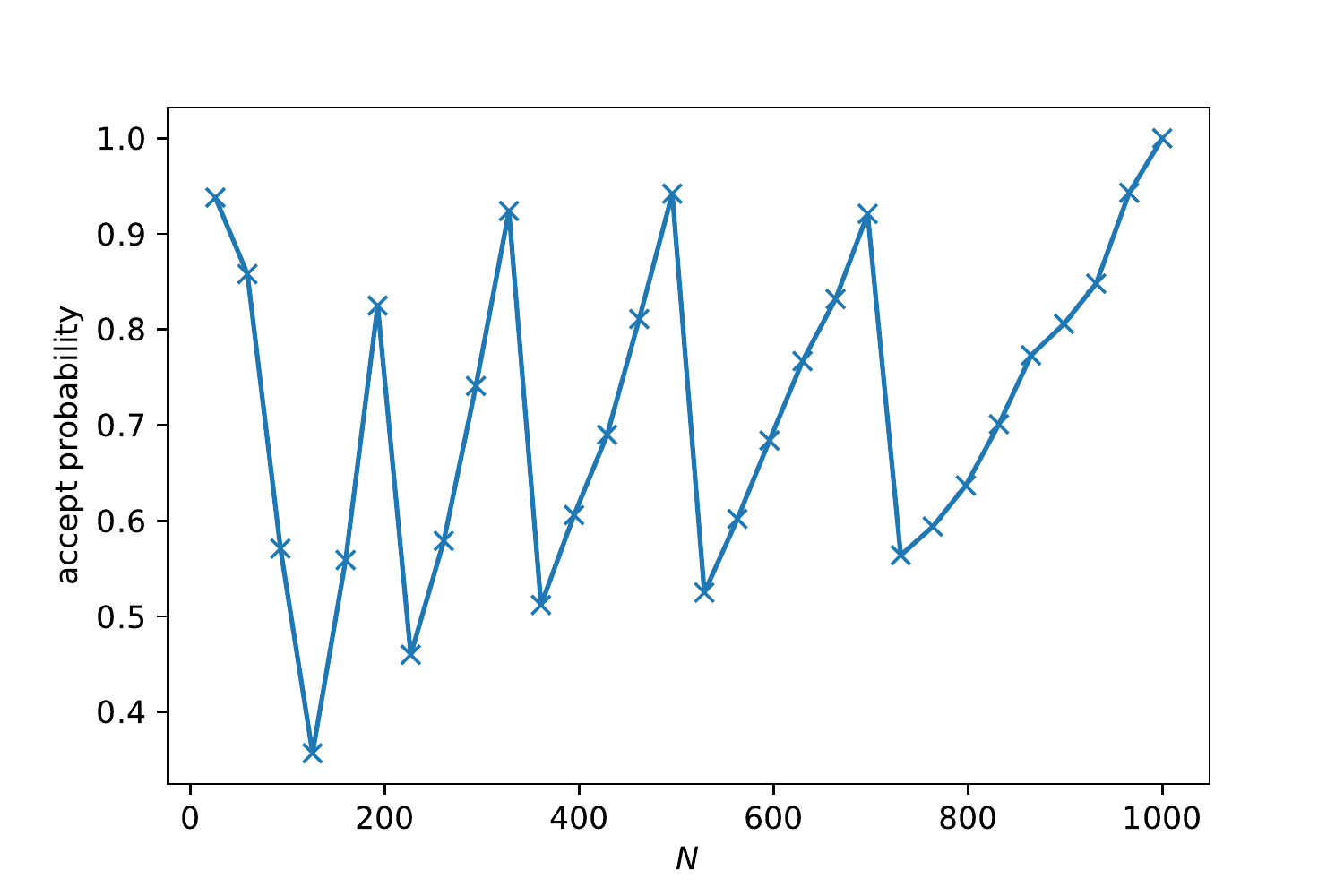}
        \caption{$d=3$}
    \end{subfigure}
    \begin{subfigure}[b]{0.3\textwidth}
        \centering
        \includegraphics[width=\textwidth]{figures/accept_prob_bootstrap_d3.pdf}
        \caption{$d=4$}
    \end{subfigure}
    \caption{Empirical acceptance probabilities for the general $N$ sampler constructed in Section \ref{subsec:general_N}. 1000 trials are used for each $N$ values.}
    \label{fig:accept_prob_general_N}
\end{figure}

\medskip
Figure \ref{fig:accept_prob_general_N} shows empirical acceptance probabilities for dimension $2, 3$ and $4$ over a range of $N$ values. 
Note that these values are just for the rejection sampler described by (\ref{eq:gen_N_reject}), treating the rejection sampler for $q$ as a black-box.
Unsurprisingly the acceptance probability has some oscillatory behaviour in $N$, with the peaks corresponding to perfect powers of $d$, followed by sharp drops as $m$ jumps from its best value of $0$ to its worst of $n^d - 1$.
That being said, the acceptance probabilities are still very good for $d=2$ above around $N=100$.
For smaller values of $N$ the acceptance probabilities are still good enough, particularly as the other components of the DPP sampler become cheaper with decreasing $N$.
$d=3, 4$ also appear feasible, though clearly less efficient that $d=2$.
Higher values of $d$ are also plausible, though the scaling with $d$ is clearly poor (c.f. \cite{NEURIPS2019_1d54c76f,bardenet2020monte} where the acceptance probability scales like $2^{-d}$.)

\begin{figure}
    \centering
    \begin{subfigure}[b]{0.3\textwidth}
        \centering
        \includegraphics[width=\textwidth]{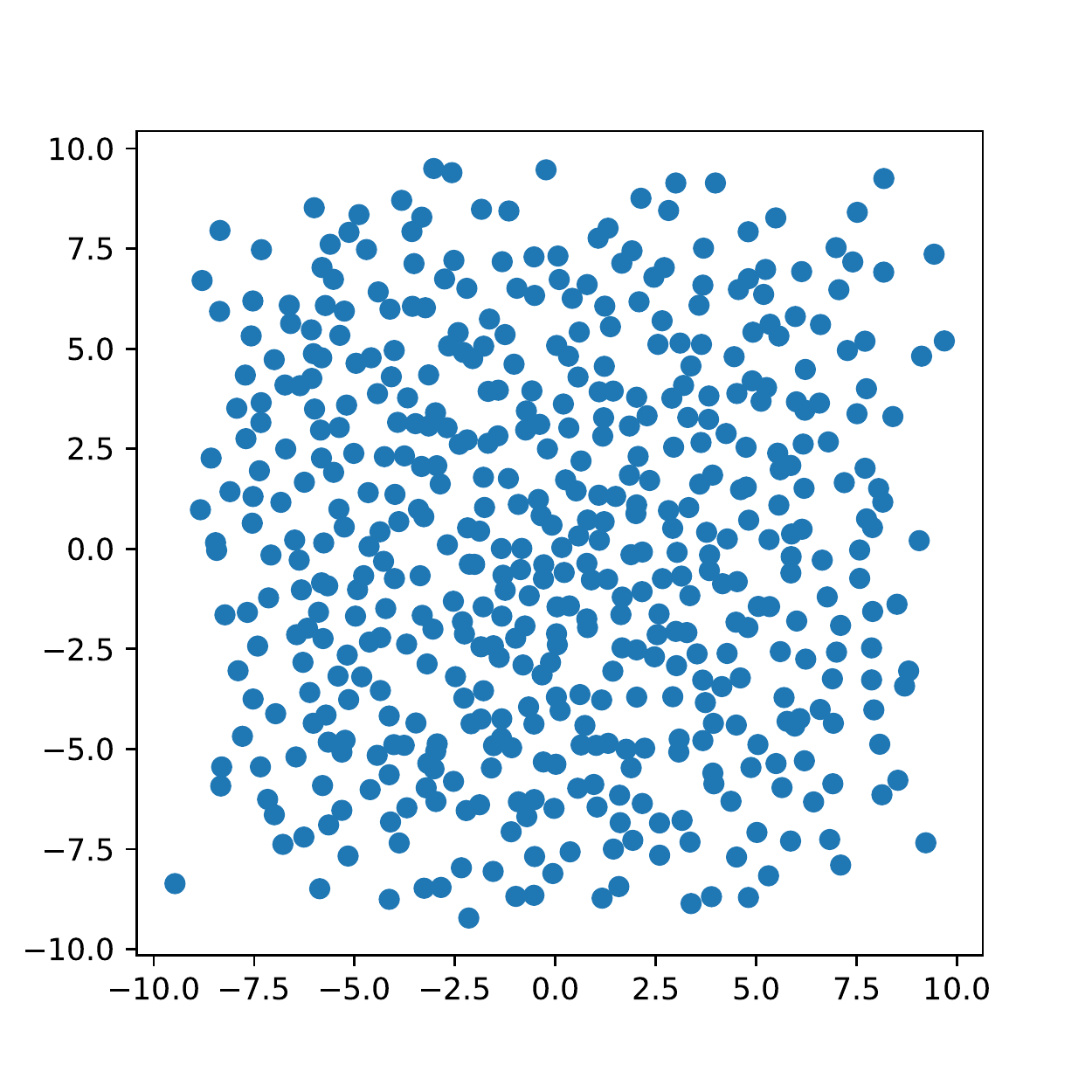}
        \caption{$\text{DPP}(\mu, K_N)$}
    \end{subfigure}
    \begin{subfigure}[b]{0.3\textwidth}
        \centering
        \includegraphics[width=\textwidth]{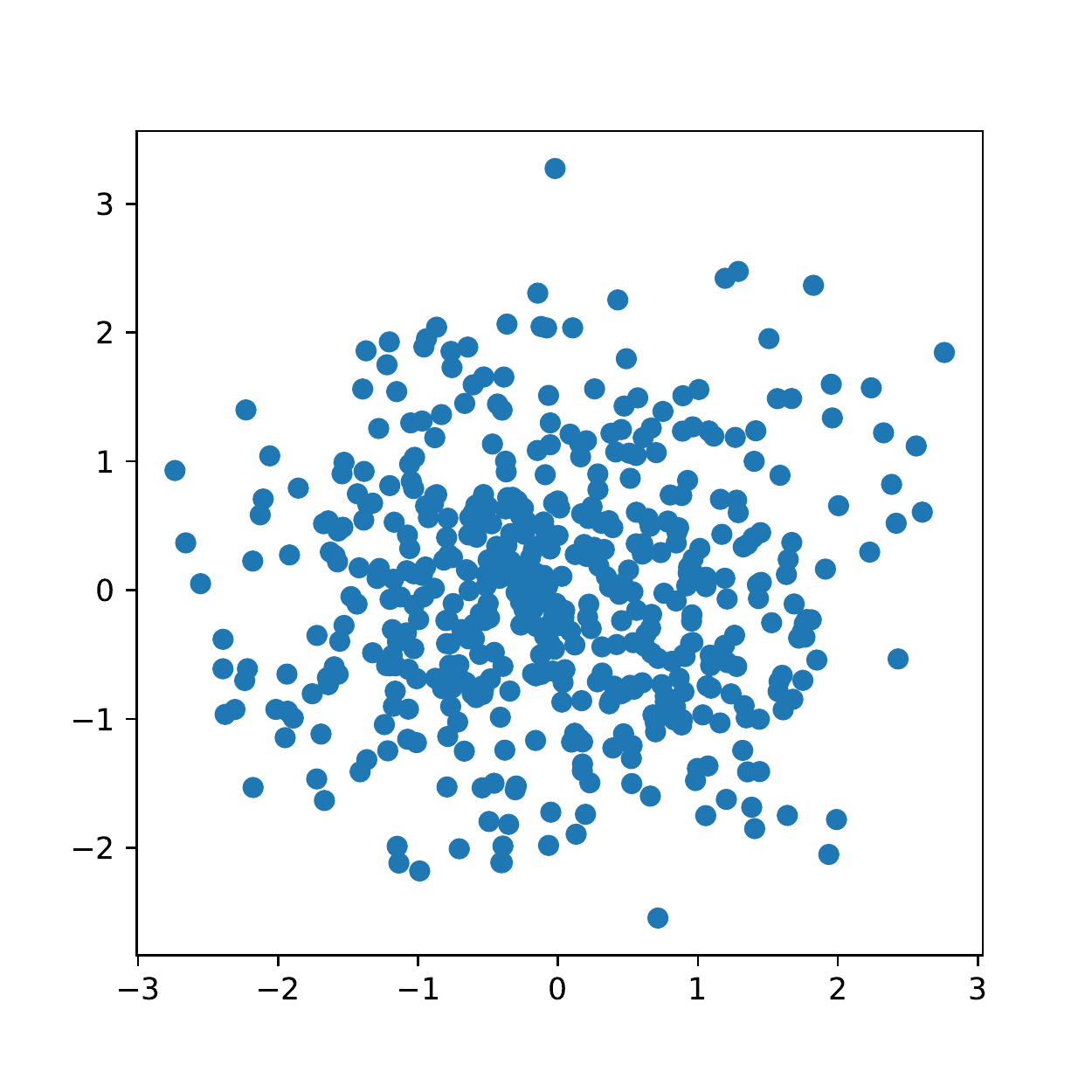}
        \caption{i.i.d. Gaussian}
    \end{subfigure}
    \begin{subfigure}[b]{0.3\textwidth}
        \centering
        \includegraphics[width=\textwidth]{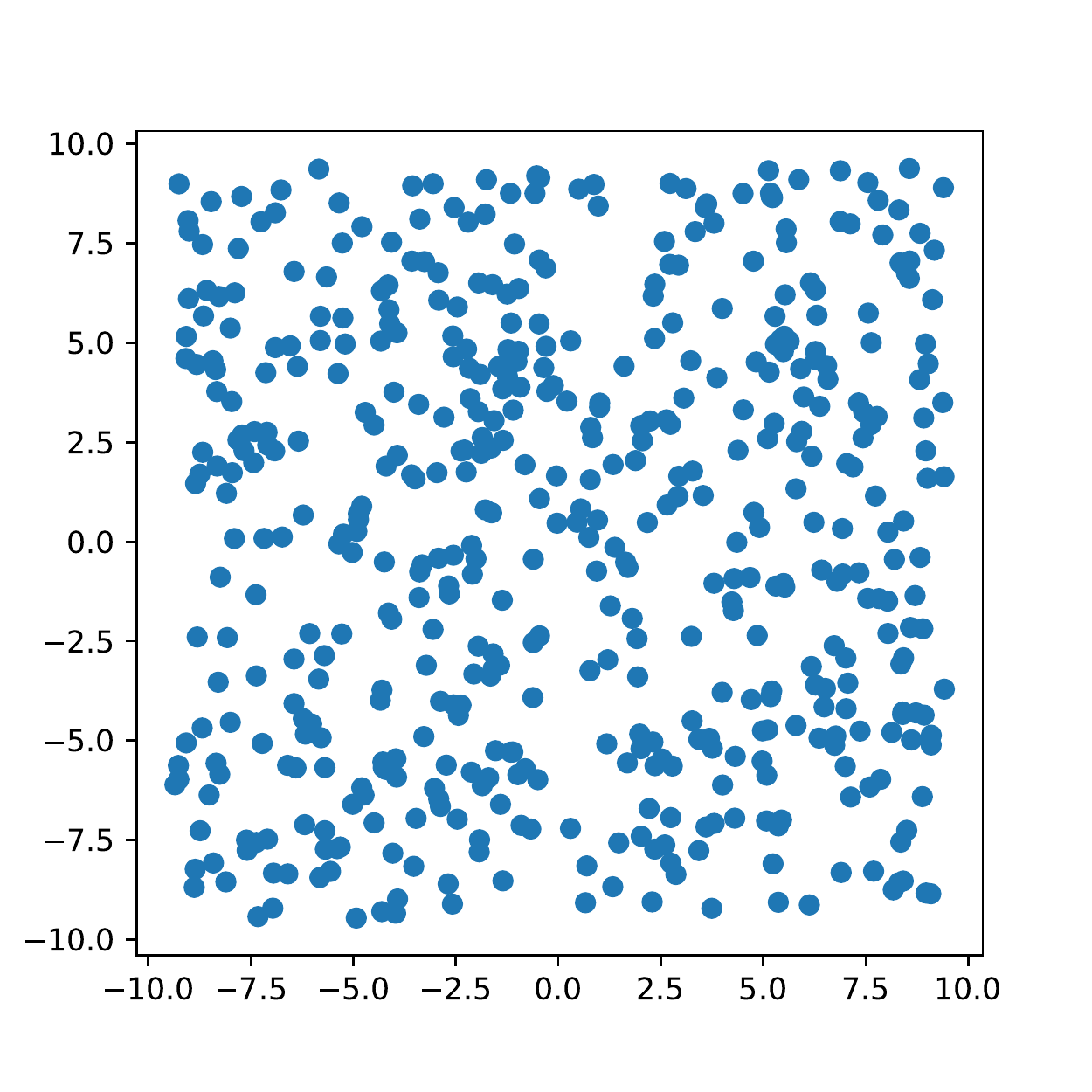}
        \caption{Poisson process}
    \end{subfigure}
    \caption{Samples in 2 dimension with $N=500$. We show the DPP and for comparison i.i.d. samples from the base Gaussian measure and also a Poisson process on $(-2N^{1/4}, 2N^{1/4})$.}
    \label{fig:2d_dpp_sample}
\end{figure}

\medskip
Figure \ref{fig:2d_dpp_sample} show a sample from the constructed Gauss-Hermite DPP in 2 dimension and with 500 points.
For comparison we also show a sample from the base measure, a standard Gaussian in 2 dimensions, and from a Poisson process on $(-2N^{1/4}, 2N^{1/4})$ (the domain in which the DPP has the vast majority of its support).
The effect of the DPP repulsion is evident when comparing it to the base measure, as the points have filled a much larger domain.
The repulsion is also evident when comparing the DPP to the Poisson process. The two processes have produced points in very similar domains, but the DPP shows much more uniformity of placement, with the Poisson process showing lots of characteristic clustering of points.

\begin{figure}
    \centering
    \includegraphics[width=0.4\textwidth]{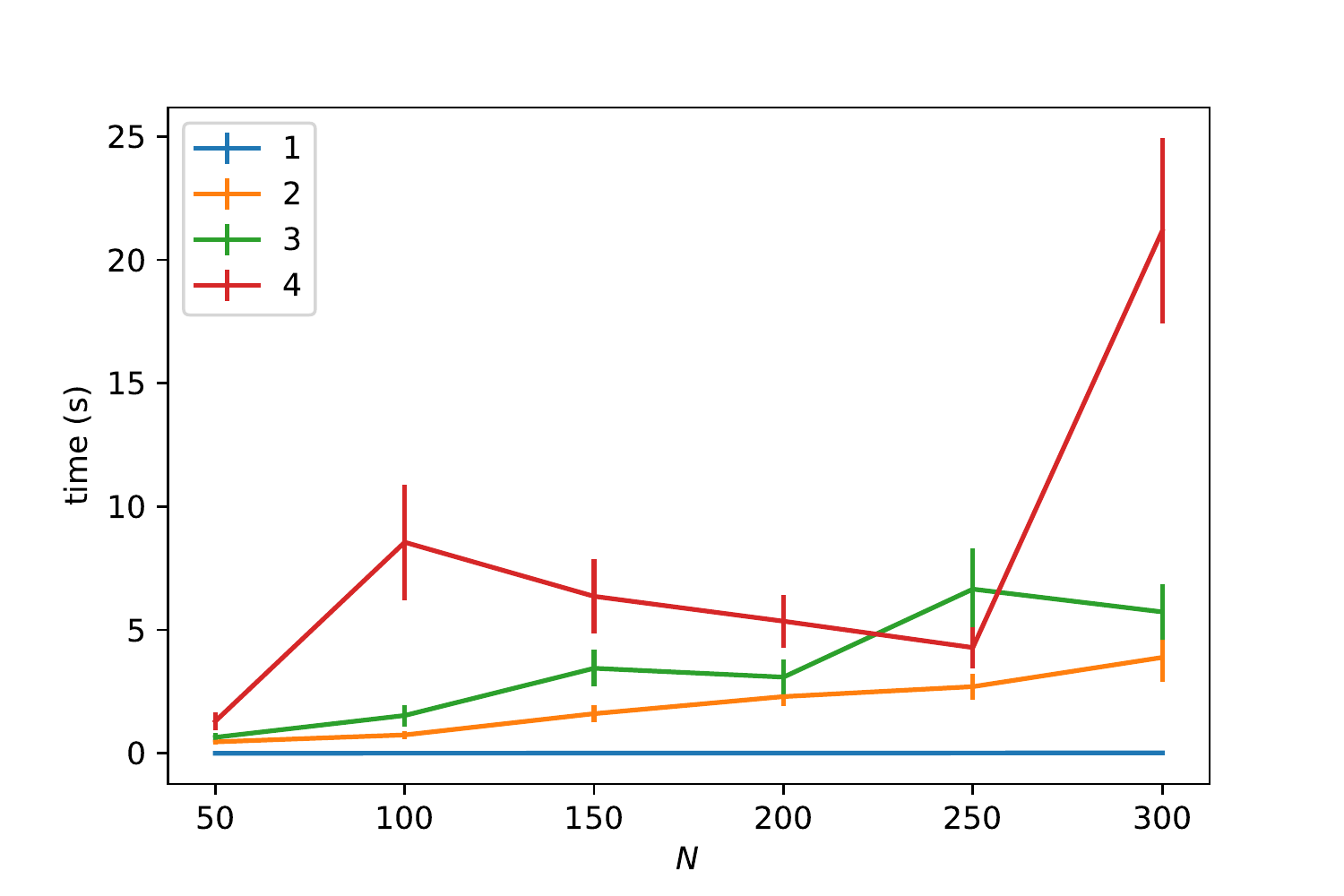}
    \caption{The wall-time of sampling from $\text{DPP}(\mu, K_N)$ for various dimensions $1,2,3,4$ and a range of $N$ values. Shown are means and one standard deviation either side over 30 repeated samplings for each $N$ and $d$.}
    \label{fig:timing}
\end{figure}

\medskip
Figure \ref{fig:timing} shows some timing results for the DPP sampler in dimensions $d=1,2,3,4$. 
The results are broadly similar to those of \cite{NEURIPS2019_1d54c76f}, which is unsurprising since the top-level chain rule step of the sampler is the same and the lower-level rejection samplers are efficient in both cases for small $d$.
Our rejection sampler for $f_N$ introduces $r_{N,d}^{-1}ds^{-1}_{\lceil N^{1/d}\rceil}$, where $s_n$ is the acceptance probability for our sampler of $\rho_{n}$, so $s_n \geq 0.7$, and $r_{N, d}$ is the acceptance probability for our overall rejection sampler for $f_N$, so it appears $r_{N,d} \geq 0.4$ for $N\geq 200$ and with most $N$ values having much better values.
Comparing the figure to the corresponding one in \cite{NEURIPS2019_1d54c76f} does suggest that for $d=4$ our rejection sampler for $f_N$ is introducing extra inefficiency beyond that of the top-level chain-rule sampler.
However, recalling the acceptance probabilities in Figure \ref{fig:accept_prob_general_N}, it appears that the poorer performance of for $d=4$ is caused by the complexity of computing many more Hermite polynomial evaluations, namely $(\lceil N^{1/d}\rceil)^d$ rather than just $N$.
This is encouraging, as the best case scenario (as achieved by \cite{NEURIPS2019_1d54c76f} in the Jacobi case) requires evaluation of the first $N^{1/d}$ Hermite polynomials. 
Our sampling strategy required $\lceil N^{1/d} \rceil$ which is of the same order of complexity for large $N$.

\section{Application to Monte Carlo integration}

In \cite{NEURIPS2019_1d54c76f} the authors present two DPP-based Monte Carlo (MC) integration schemes (BH and EZ) for integrals of the from
\begin{align}
    \int_{[-1, 1]^d} d\mu(\x) f(\x)
\end{align}
where $\mu$ is multivariate Jacobi measure on $[-1, 1]^d$.
Armed with our multivariate Gauss-Hermite DPP, we can also apply their methods to integrals of the form
\begin{align}
    \int_{\R^d} d\mu(\x) f(\x)
\end{align}
where recall $d\mu(\x) = e^{-\frac{\x^2}{2}}d\x$. Moreover, given some mean $\vec{\mu}$ and covariance matrix $\vec{\Sigma}$ we have
\begin{align}
    \int_{\R^d}d\x ~ \mathcal{N}(\x; \vec{\mu}, \vec{\Sigma}) f(\x) &= \frac{1}{(2\pi)^{d/2}}\int_{\R^d}d\mu(\x)f(\vec{\Sigma}^{1/2}\x + \vec{\mu})
\end{align}
and so estimation of integrals against \emph{any Gaussian probability measure} is possible by the reparametrisation trick.
Other than switching to our Gauss-Hermite DPP from the Jacobi DPP, we make no other changes to the BH and EZ approaches.
% TODO: note rescaling of measure for DPP.
\subsection{Perfect esimation of polynomials}
Consider functions $\R^d\rightarrow\R$ of the form \begin{align}
    f(x) = \sum_{i_1, \ldots, i_d=1}^p a_{\vec{i}} \prod_{k=1}^d x_k^{i_k}
\end{align}
where $a_{\vec{i}}$ are coefficients. The integral of $f$ against Gaussian measure is simple to compute:
\begin{align}\label{eq:poly_true}
    \int_{\R^d} d\x ~ \mathcal{N}(\x; \vec{\mu}, \vec{\Sigma}) f(\x) = \sum_{i_1, \ldots, i_d=1}^p a_{\vec{i}}  \prod_{k=1}^d \int_{\R} dx ~ \frac{e^{-\frac{x^2}{2}}}{\sqrt{2\pi}} x_k^{i_k} = \sum_{i_1, \ldots, i_d=1}^p a_{\vec{i}}  \prod_{k=1}^d m(i_k)
\end{align}
where $m(i) = (i-1)!!$ if $i$ even and $m(i) = 0$ otherwise.

\begin{figure}
    \centering
    \begin{subfigure}[b]{0.3\textwidth}
        \centering
        \includegraphics[width=\textwidth]{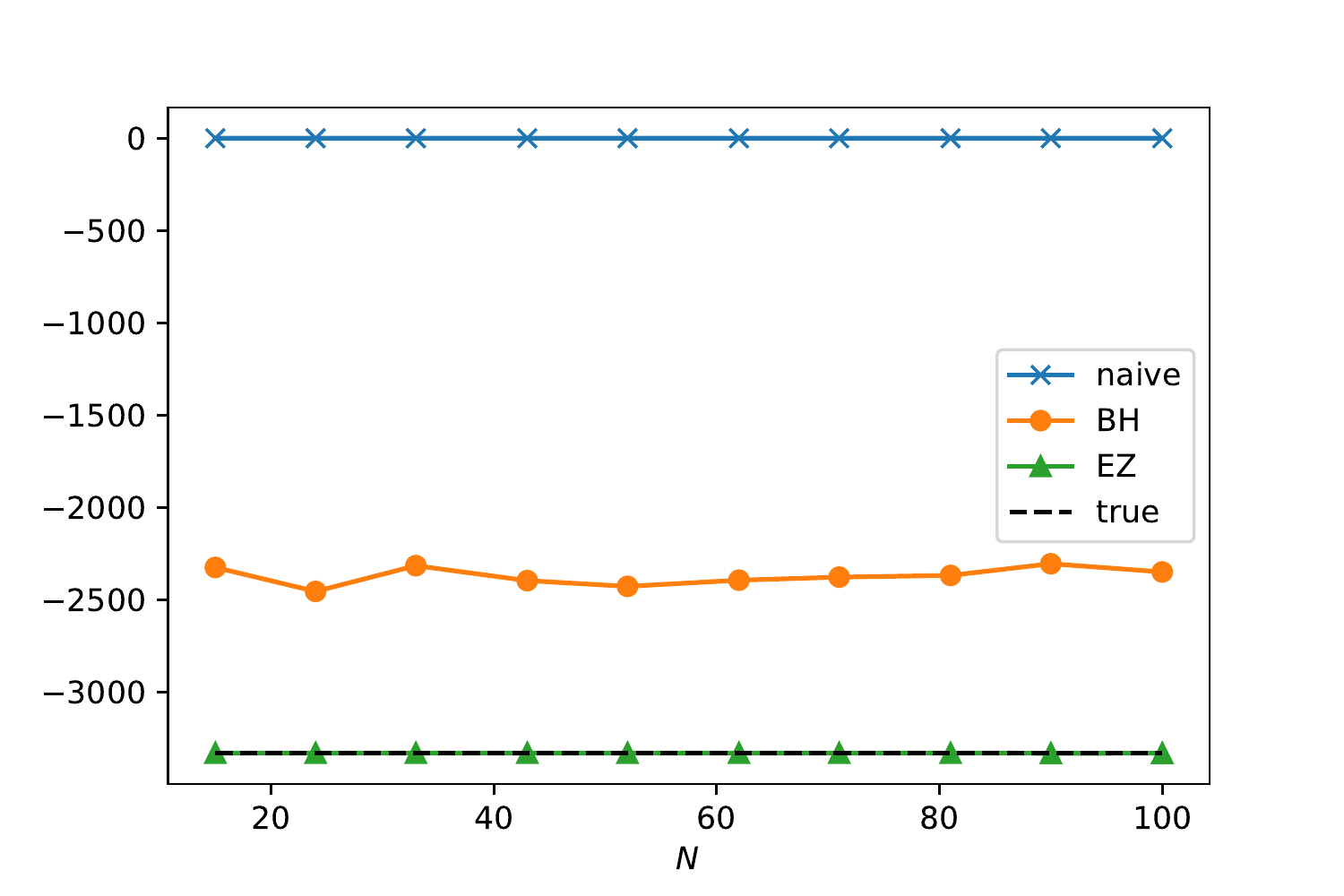}
    \end{subfigure}
    \begin{subfigure}[b]{0.3\textwidth}
        \centering
        \includegraphics[width=\textwidth]{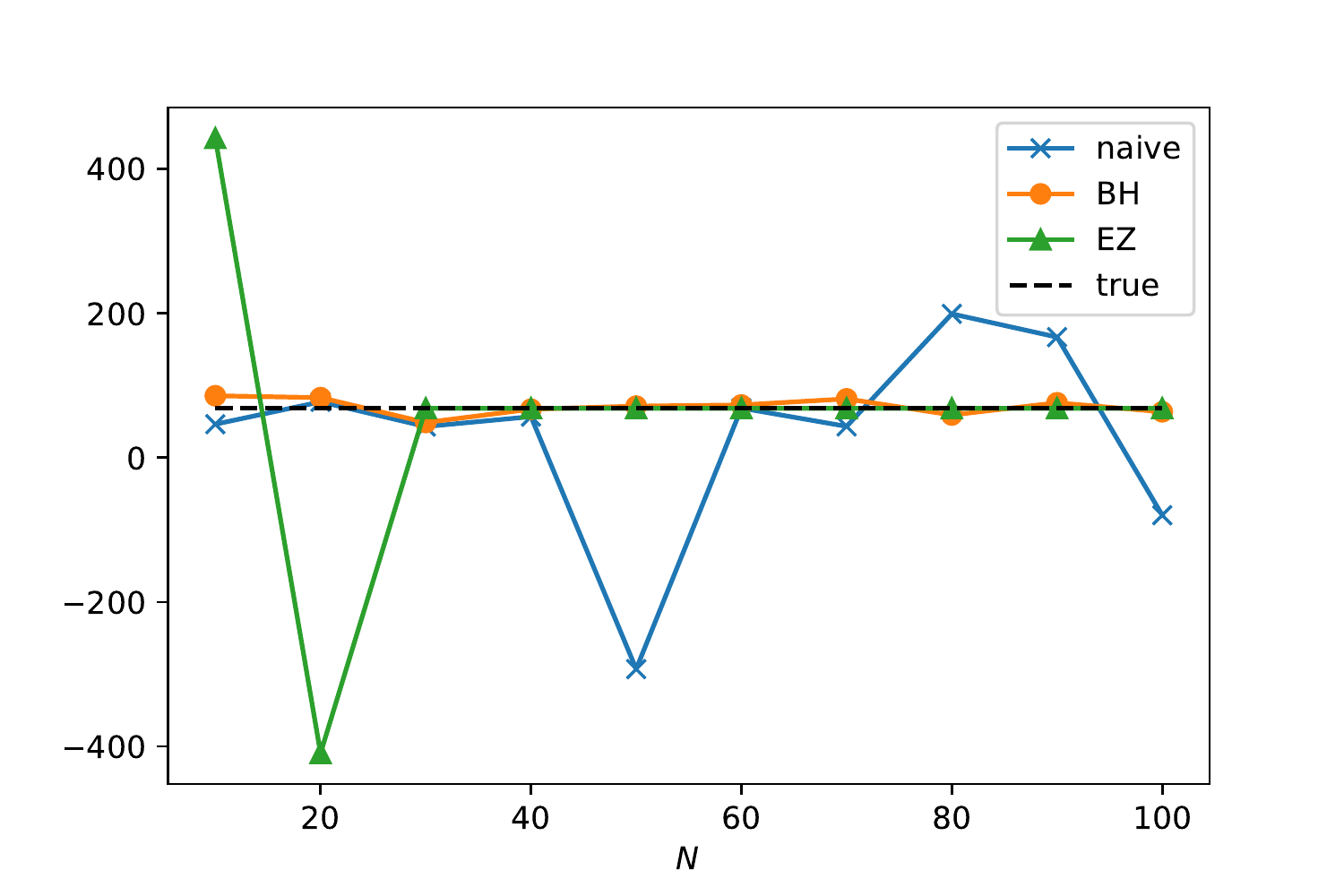}
    \end{subfigure}
    \begin{subfigure}[b]{0.3\textwidth}
        \centering
        \includegraphics[width=\textwidth]{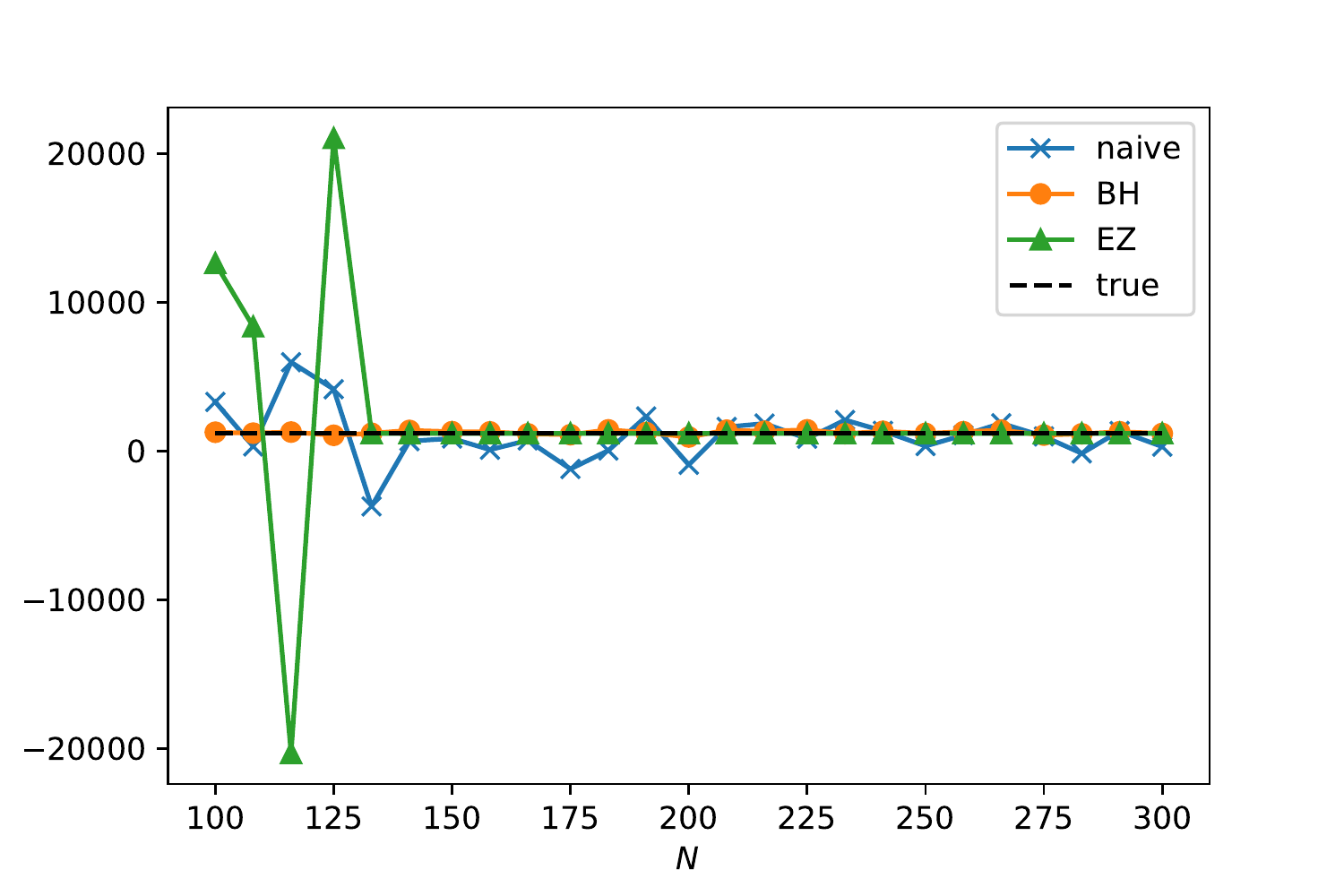}
    \end{subfigure}
    \begin{subfigure}[b]{0.3\textwidth}
        \centering
        \includegraphics[width=\textwidth]{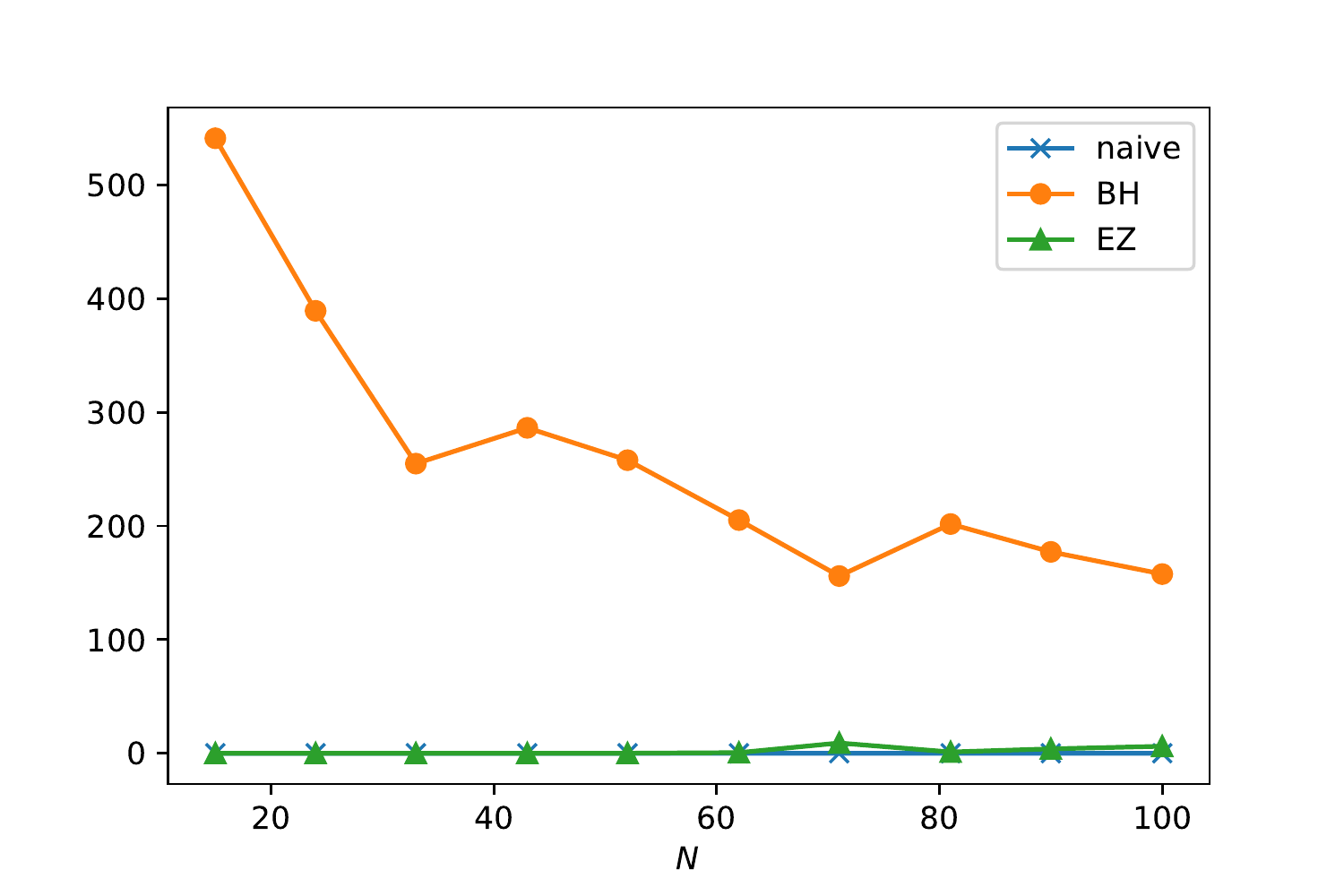}
        \caption{$d=1$}
    \end{subfigure}
    \begin{subfigure}[b]{0.3\textwidth}
        \centering
        \includegraphics[width=\textwidth]{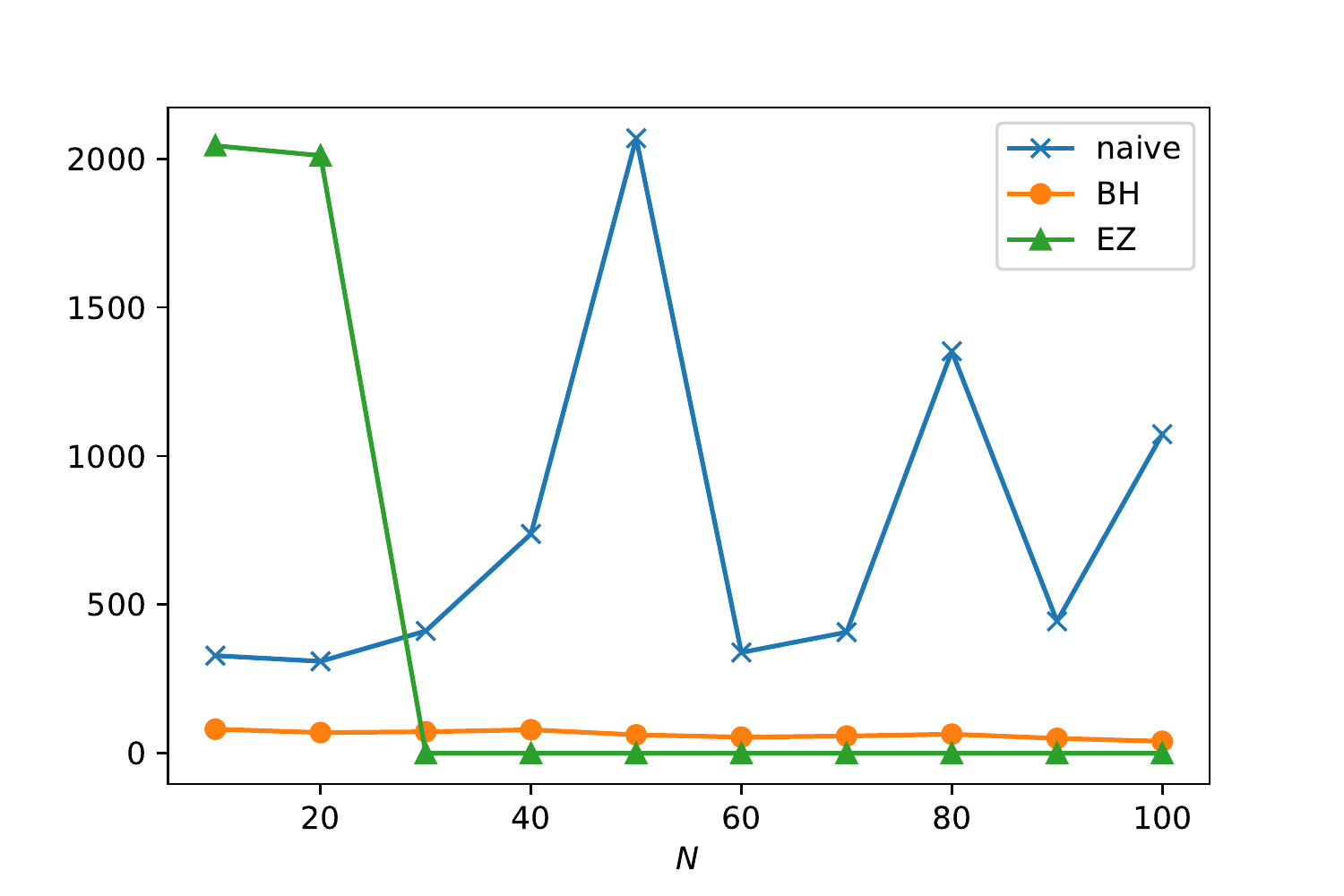}
        \caption{$d=2$}
    \end{subfigure}
    \begin{subfigure}[b]{0.3\textwidth}
        \centering
        \includegraphics[width=\textwidth]{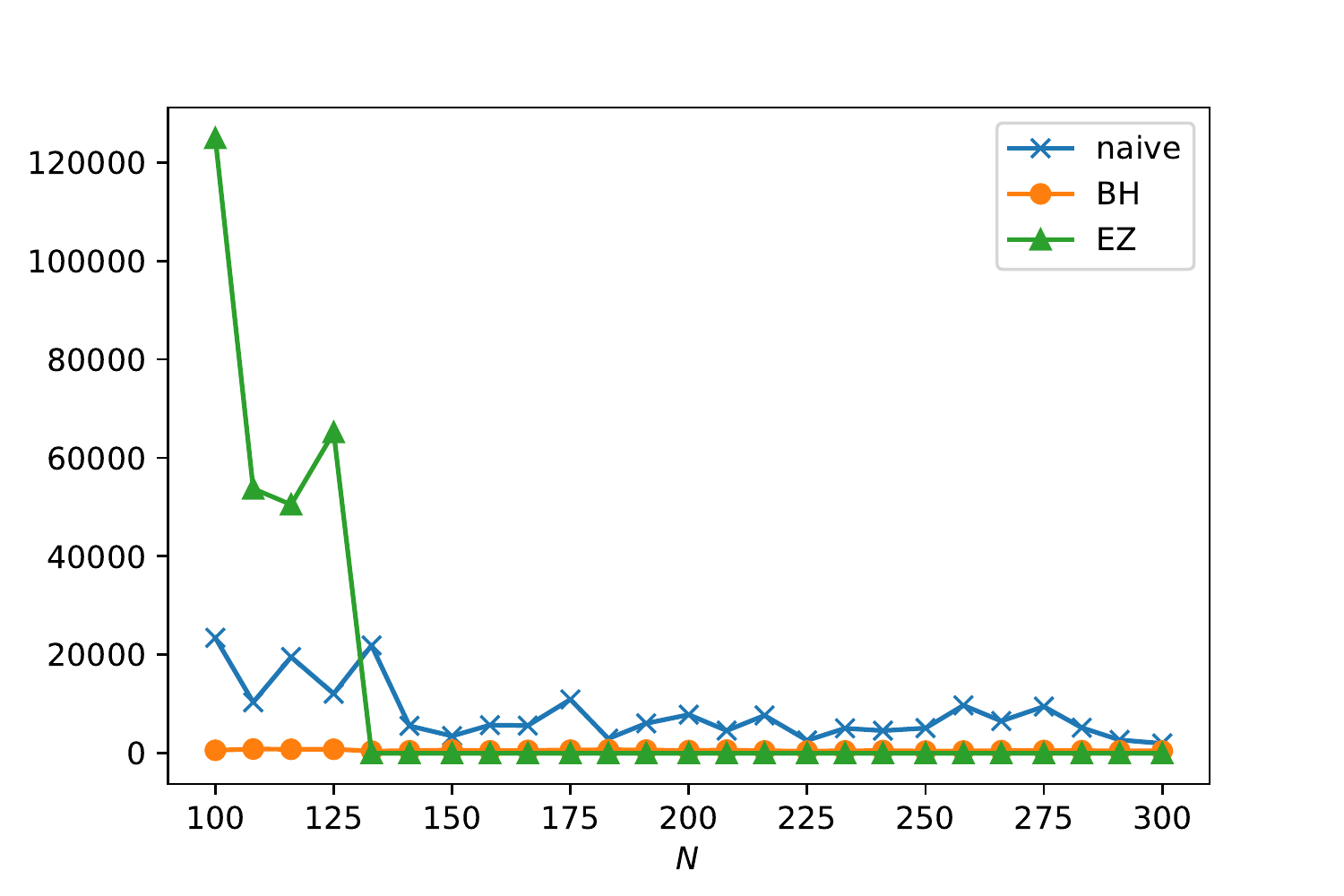}
        \caption{$d=3$}
    \end{subfigure}
    \caption{Result of MC integration on multivariate polynomials using Gauss-Hermite DPPs and comparing the na\"{i}ve MC approach to the BH and EZ approaches. The top row shows the sample means over 30 repetitions and the bottom row shows the sample standard deviation.}
    \label{fig:poly_results}
\end{figure}

Figure \ref{fig:poly_results} shows the results of using the BH, EZ and na\"{i}ve MC integration approaches.
We show the mean and standard deviation over 30 repeated samples. The polynomial are selected at random but are kept fixed throughout.
The degree of the polynomials is 10 for $d=1$, 5 for $d=2, 3$.
The reduction in degree for higher dimension is necessary to prevent the true values given by (\ref{eq:poly_true}) from becoming unreasonably large.

The results generally accord with those of \cite{NEURIPS2019_1d54c76f}, with the the EZ estimator performing best but become less reliable with increasing $d$, in particular prone to ill-conditioning leading to enormous outliers.
The BH estimator becomes more favourable with increasing dimension.
All estimators benefit from increasing sample size $N$, but the EZ estimator benefits most dramatically.
For example in the $d=2$ case the EZ estimator is very poor until around $N=30$ when, beyond which it recovers the true value with zero variance (perfect estimation).
This can be explained as follows: we chose a degree $5$ polynomial. For $N < 25$, the kernel $K_N$ includes only multivariate Hermite polynomials of at most degree $4$, hence the chosen polynomial is not in the span of the RKHS basis.

\subsection{Marginalised Gaussian process posteriors}\label{subsec:gp}
\begin{figure}
    \centering
    \begin{subfigure}[b]{0.4\textwidth}
        \centering
        \includegraphics[width=\textwidth]{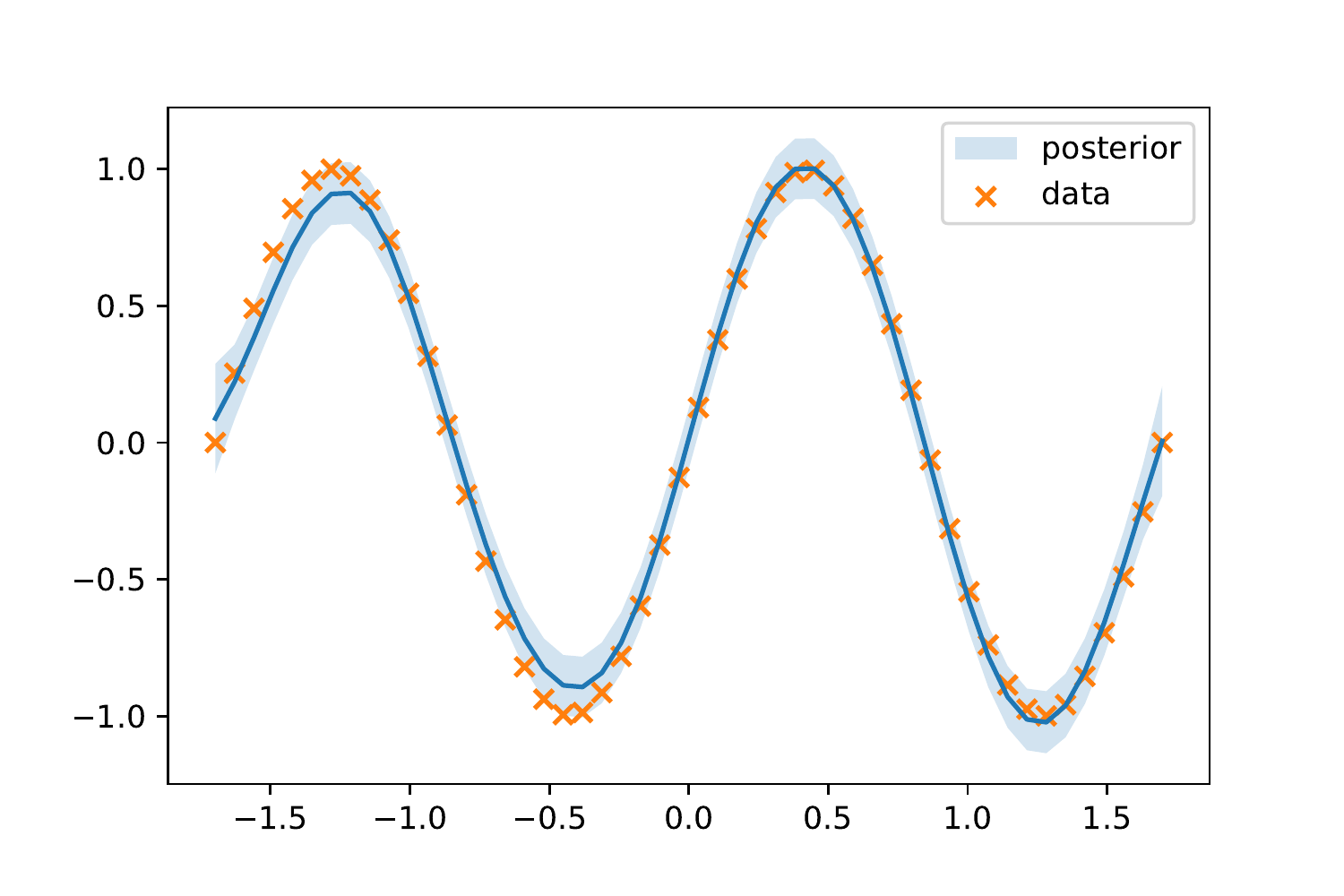}
        \caption{Point estimate.}
        \label{fig:gp_posterior_pnt}
    \end{subfigure}
    \begin{subfigure}[b]{0.4\textwidth}
        \centering
        \includegraphics[width=\textwidth]{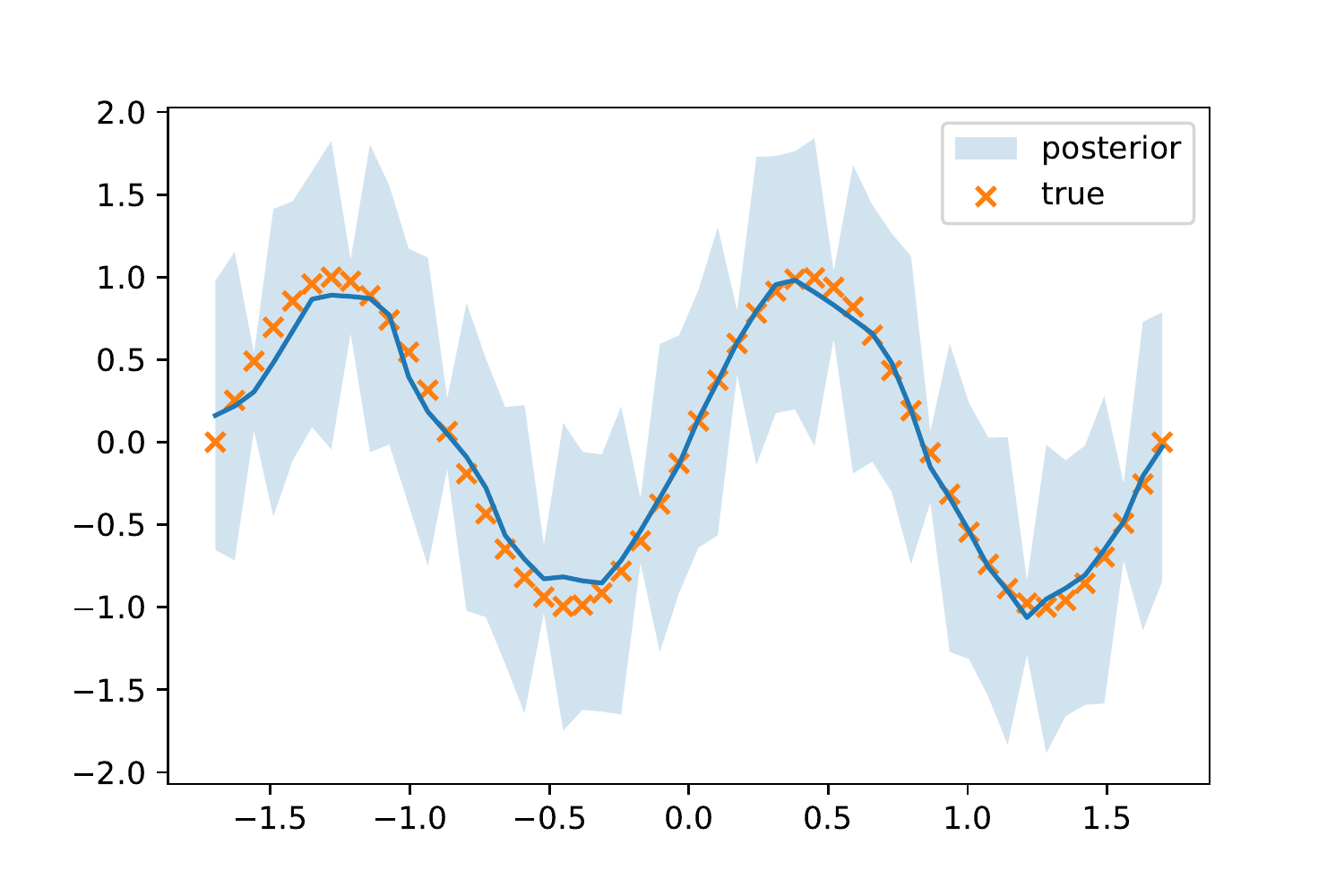}
        \caption{EZ marginalised.}
    \end{subfigure}
    \caption{Marginalised GP posterior using EZ MC integration to compute $\bar{\mu}$ and $\bar{\sigma}$. The shaded region in (b) shows $(\bar{\mu} - 1.96\bar{\sigma}, \bar{\mu}+1.96\bar{\sigma})$, whereas the shaded region in (a) is the standard posterior $95\%$ confindence interval for the point estimate GP.}
    \label{fig:gp_posterior}
\end{figure}

Here we will consider a Gaussian process (GP) posterior on a simple synthetic dataset.
The synthetic data are just draw from a one dimensional sine wave and are shown in Figure \ref{fig:gp_posterior_pnt}.
We will use a simple scaled radial basis function (RBF) kernel without automatic relevance determination:

$$ k(x, x') = v \exp\left(-\frac{(x - x)^2}{l^2}\right)$$

where $v>0$ and $l>0$ are hyperparameters.
In standard GP practice, the hyperparameters $v,l$ are optimised as point estimates by maximising the log marginal likelihood of the training data under the model.
To enforce the positivity constrains on $v,l$ we adopt the standard practice (used e.g. by GPyTorch \cite{gardner2018gpytorch}) of writing them as softplus transforms of raw hyperparameters:

\begin{align}
    v = \log(\exp(\theta_1) + 1), ~~~l = \log(\exp(\theta_2) + 1).
\end{align}

In Bayesian treatments of GP hyperparameters, one promotes a point estimate of $\theta$ to a full posterior distribution (see e.g. \cite{lalchand2020approximate}).
Further pursuit of such ideas is well outside the scope of this paper, however it suffices to note that a Gaussian form of the posterior is the most practical choice (e.g. as in Variational inference approaches \cite{lalchand2020approximate}).
We will place a Gaussian over the raw hyperparameters $\theta$.
For demonstration purposes, we will take a unit variance factorised Gaussian centred on the point-estimate raw-hyperparameters from applying marginal likelihood maximisation.\footnote{This is not dissimilar to a Laplace approximation of a posterior over the hyperparameters, though the variance is certainly not correct.}
We will consider a simple 1-dimensional index space and a simple single-output regression problem on synthetic data.

\medskip
We now seek the hyperparameter-marginalised GP posterior confidence intervals. 
The point-wise CDF is

\begin{align}\label{eq:psi_cdf}
    \Psi(y) = \int d\theta ~ \Phi\left(\frac{y - \mu(\theta + \theta_*)}{\sigma(\theta + \theta_*)}\right) \frac{e^{-\frac{\theta^2}{2}}}{2\pi}
\end{align}
where $\theta_*$ is the centre of the raw-hyperparameter Gaussian obtained from MLL maximisation, $\mu,\sigma$ are the point-wise posterior mean and standard deviation of the GP posterior as a function of $\theta$, and $\Phi$ is the standard Gaussian CDF.
There is an implicit dependence throughout on the index point $x$.
We then seek $y_2 = \Psi^{-1}(0.975), ~ y_1 = \Psi^{-1}(0.025)$, the boundaries of a confidence interval.
Rather than embarking on numerical inversion of $\Psi$, we will opt for simpler heuristics that nevertheless demonstrate the DPP MC integration methods, which is our objective.
We compute 

\begin{align}
    \bar{\mu} = \int d\theta ~ \frac{e^{-\frac{\theta^2}{2}}}{2\pi}  \mu(\theta + \theta_*)
\end{align}

and 
\begin{align}
    \bar{\sigma} = \int d\theta ~ \frac{e^{-\frac{\theta^2}{2}}}{2\pi}  \sigma(\theta + \theta_*).
\end{align}
These yield confidence intervals $(\bar{\mu} - 1.96\bar{\sigma}, \bar{\mu}+1.96\bar{\sigma})$, which are certainly not the posterior $95\%$ confidence intervals of the marginalised GP, but are perhaps not dissimilar in some cases.
Figure \ref{fig:gp_posterior} shows the point estimate posterior on a test set alongside the same EZ marginalised pseudo posterior.
We note merely that the increased posterior uncertainty appears reasonable and also that the BH and na\"{i}ve plots are very similar, but with slightly higher variance in the na\"{i}ve case.

\begin{figure}
    \centering
    \begin{subfigure}[b]{0.24\textwidth}
        \centering
        \includegraphics[width=\textwidth]{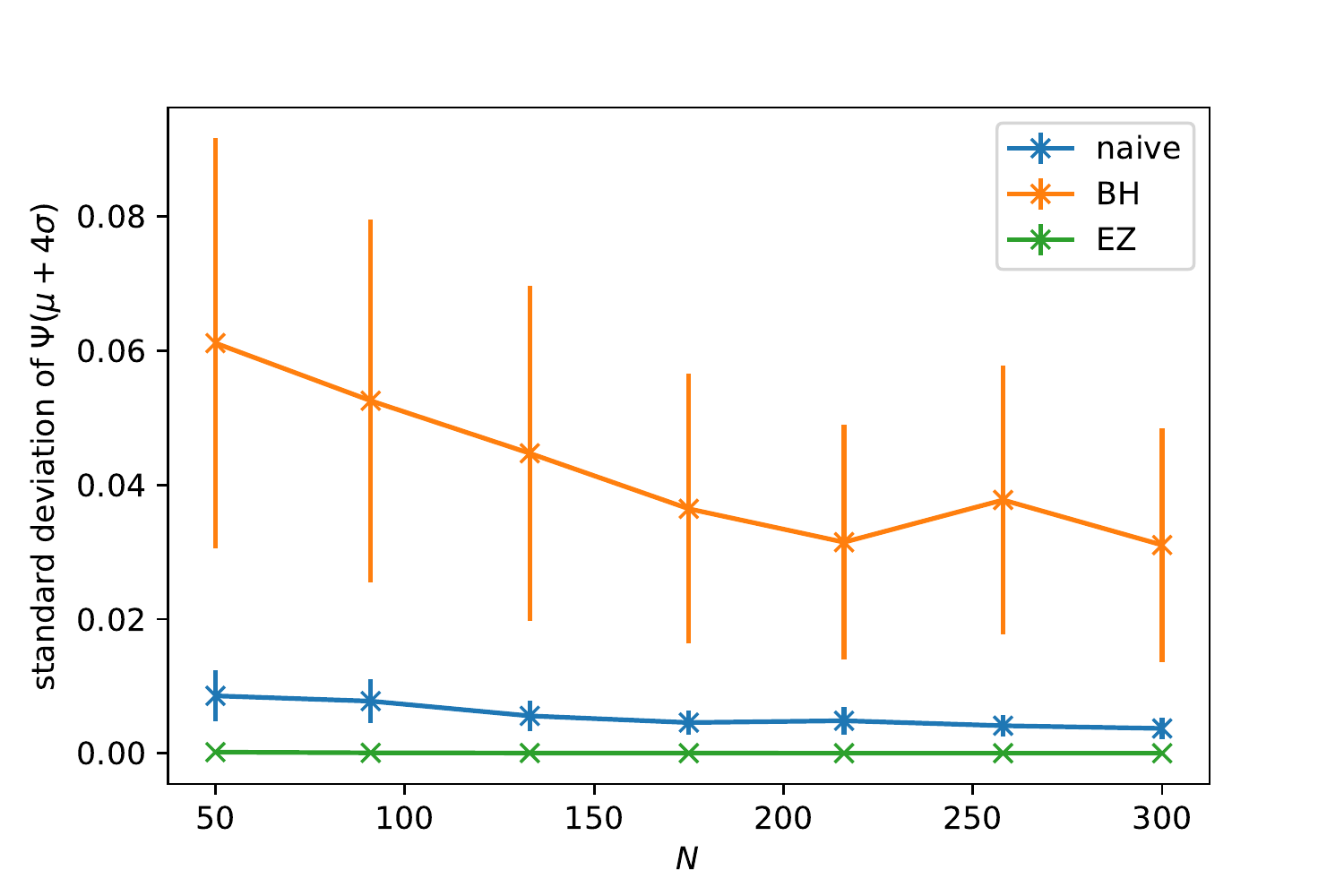}
        \caption{$y=\bar{\mu} - 4\bar{\sigma}$}
    \end{subfigure}
    \begin{subfigure}[b]{0.24\textwidth}
        \centering
        \includegraphics[width=\textwidth]{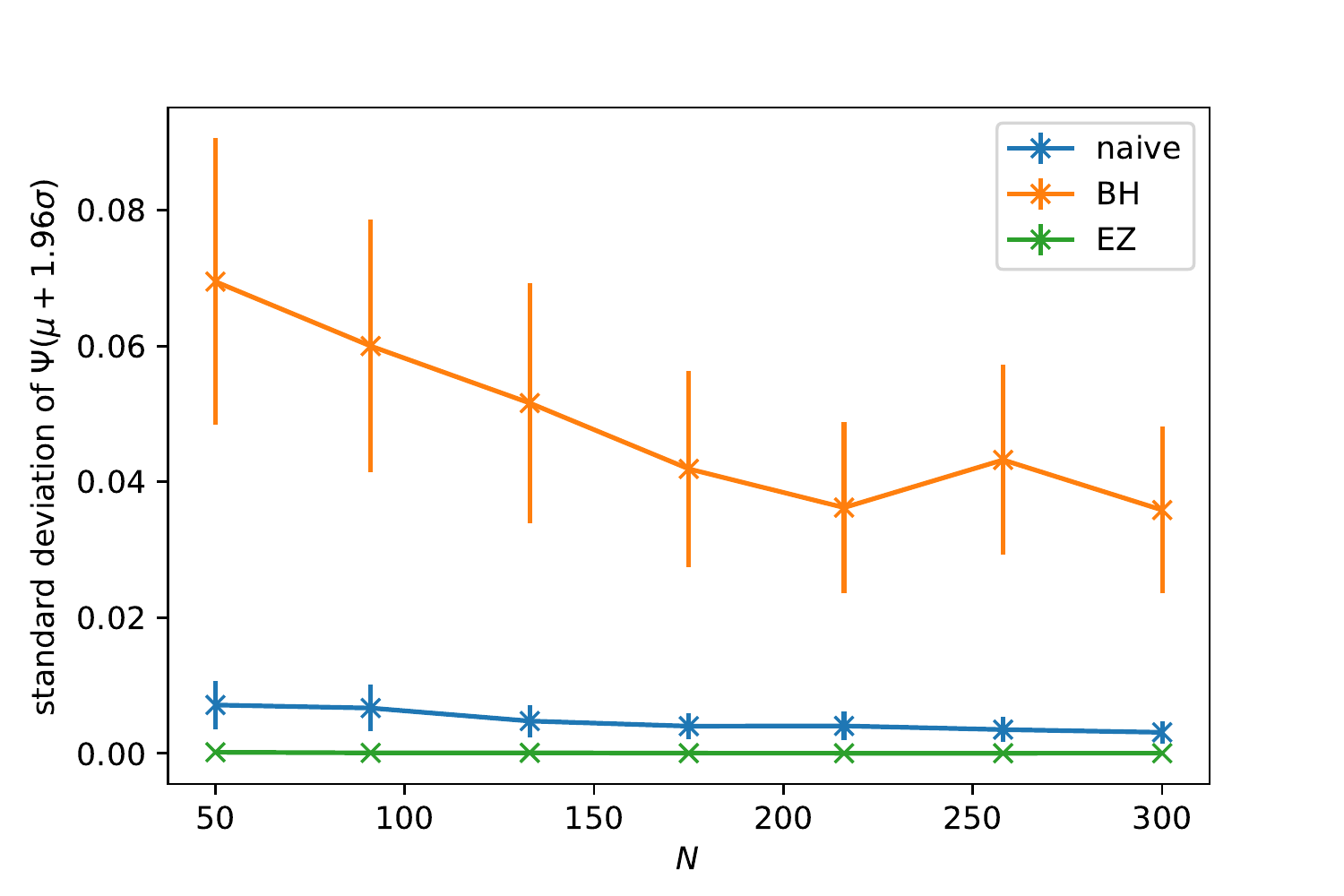}
        \caption{$y=\bar{\mu} - 1.96\bar{\sigma}$}
    \end{subfigure}
    \begin{subfigure}[b]{0.24\textwidth}
        \centering
        \includegraphics[width=\textwidth]{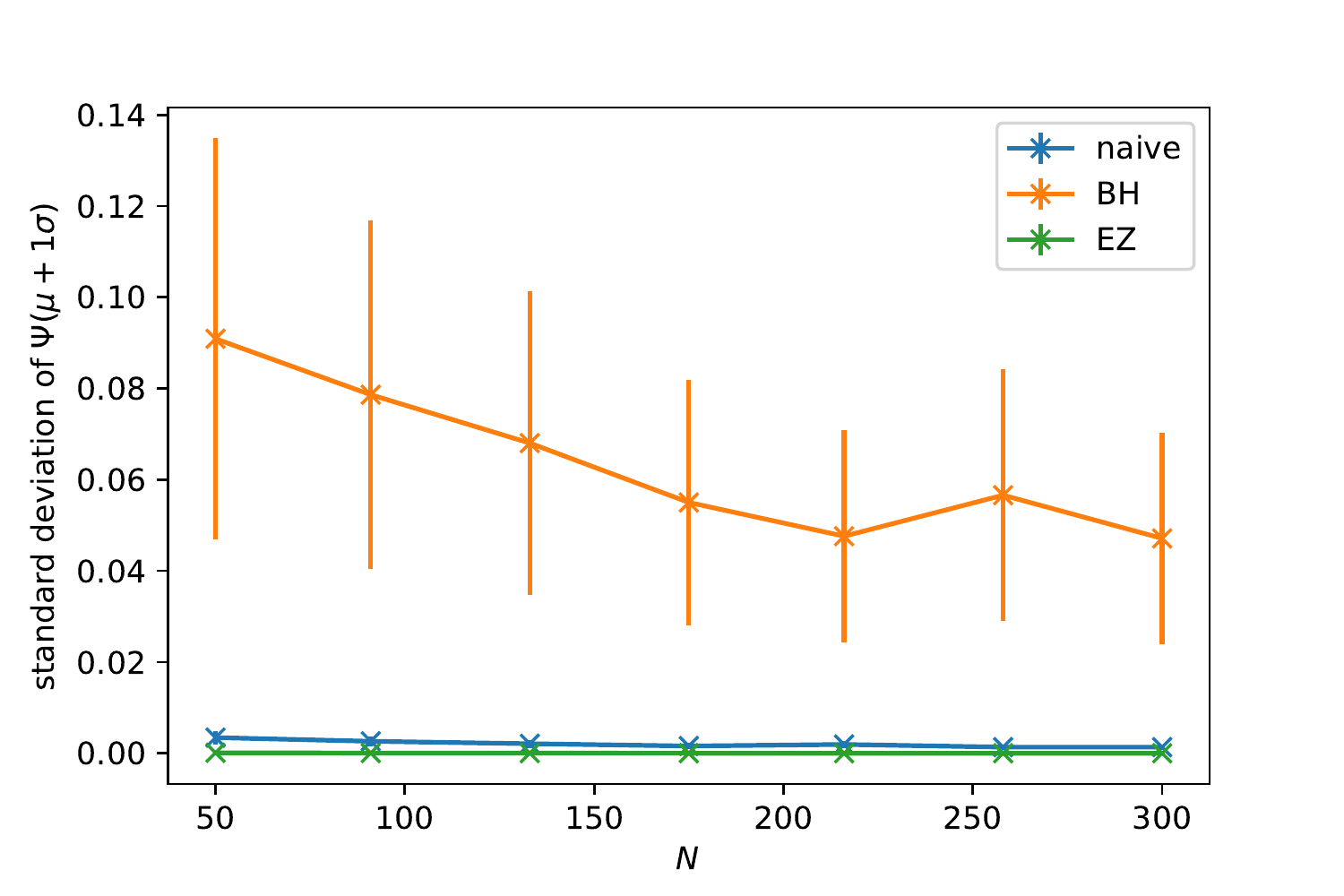}
        \caption{$y=\bar{\mu} - \bar{\sigma}$}
    \end{subfigure}
    \begin{subfigure}[b]{0.24\textwidth}
        \centering
        \includegraphics[width=\textwidth]{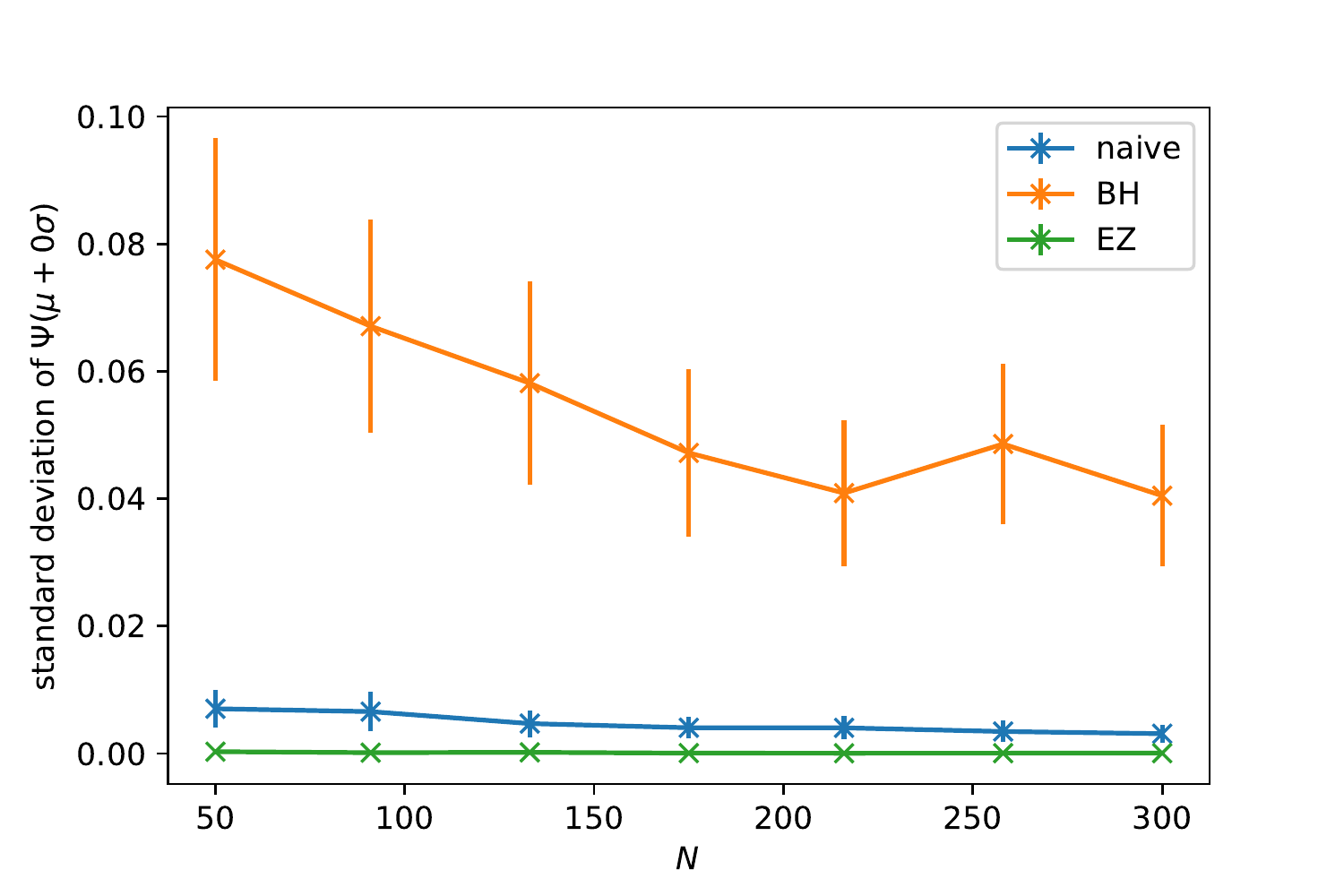}
        \caption{$y=\bar{\mu}$}
    \end{subfigure}
    \begin{subfigure}[b]{0.24\textwidth}
        \centering
        \includegraphics[width=\textwidth]{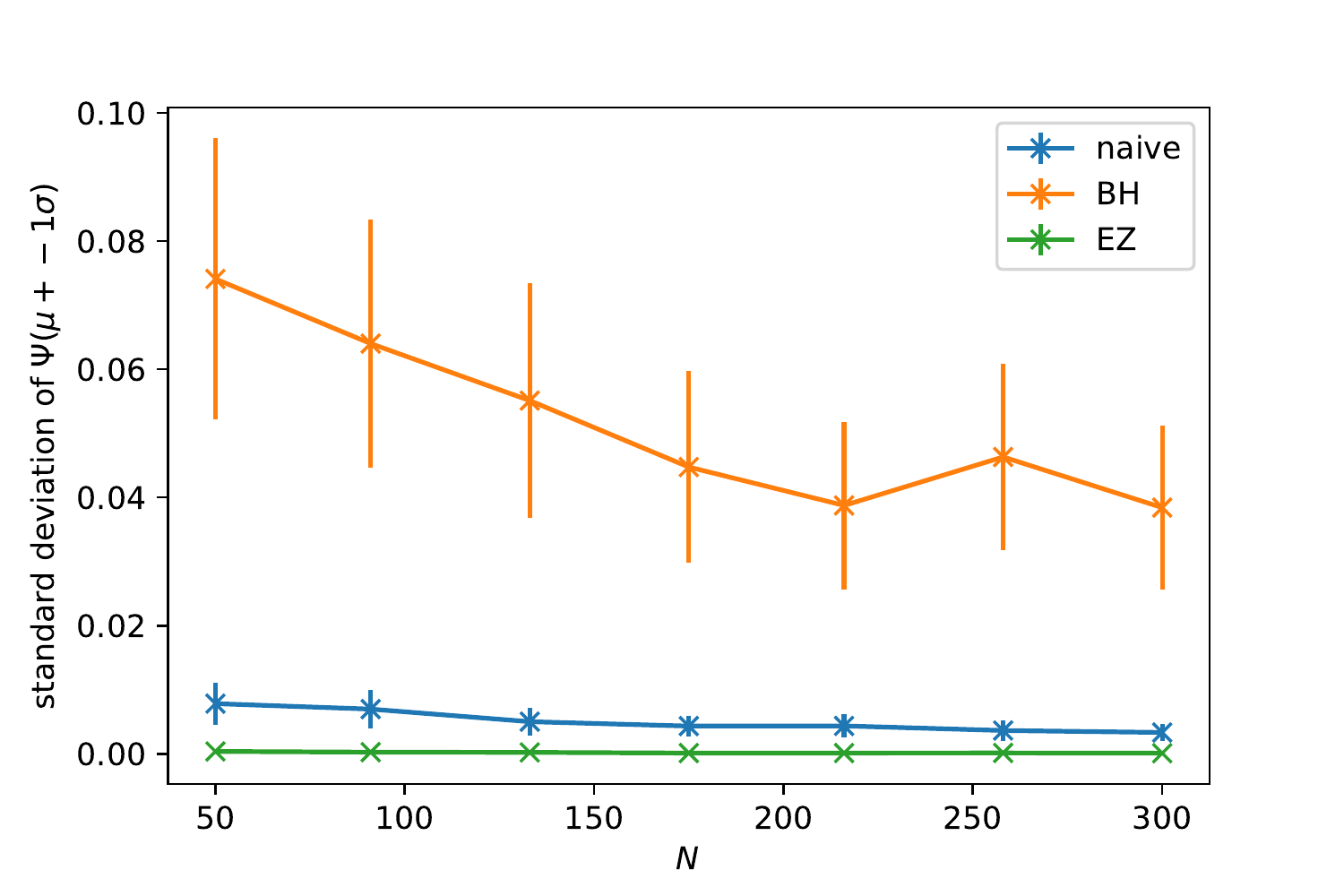}
        \caption{$y=\bar{\mu} + \bar{\sigma}$}
    \end{subfigure}
    \begin{subfigure}[b]{0.24\textwidth}
        \centering
        \includegraphics[width=\textwidth]{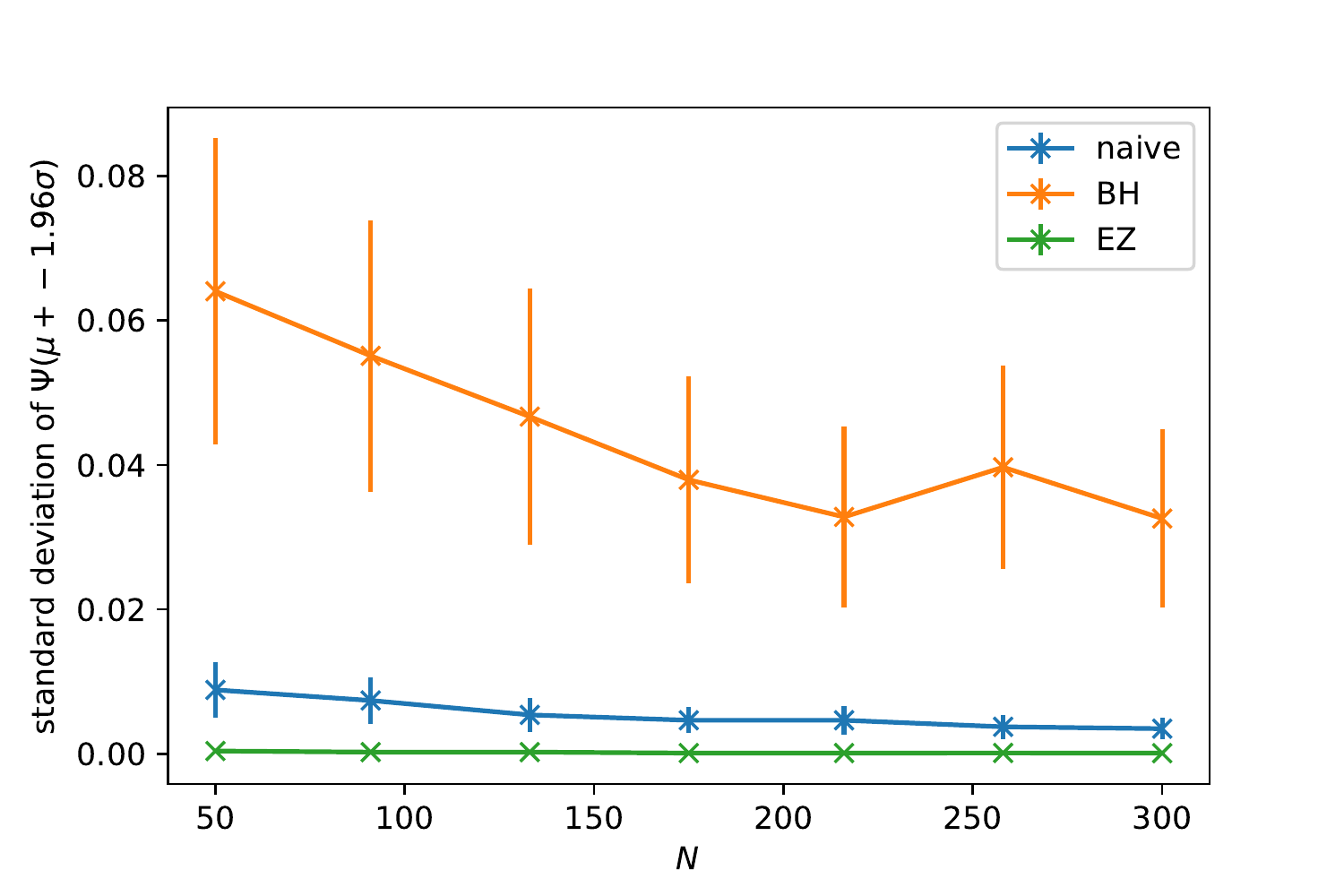}
        \caption{$y=\bar{\mu} + 1.96\bar{\sigma}$}
    \end{subfigure}
    \begin{subfigure}[b]{0.24\textwidth}
        \centering
        \includegraphics[width=\textwidth]{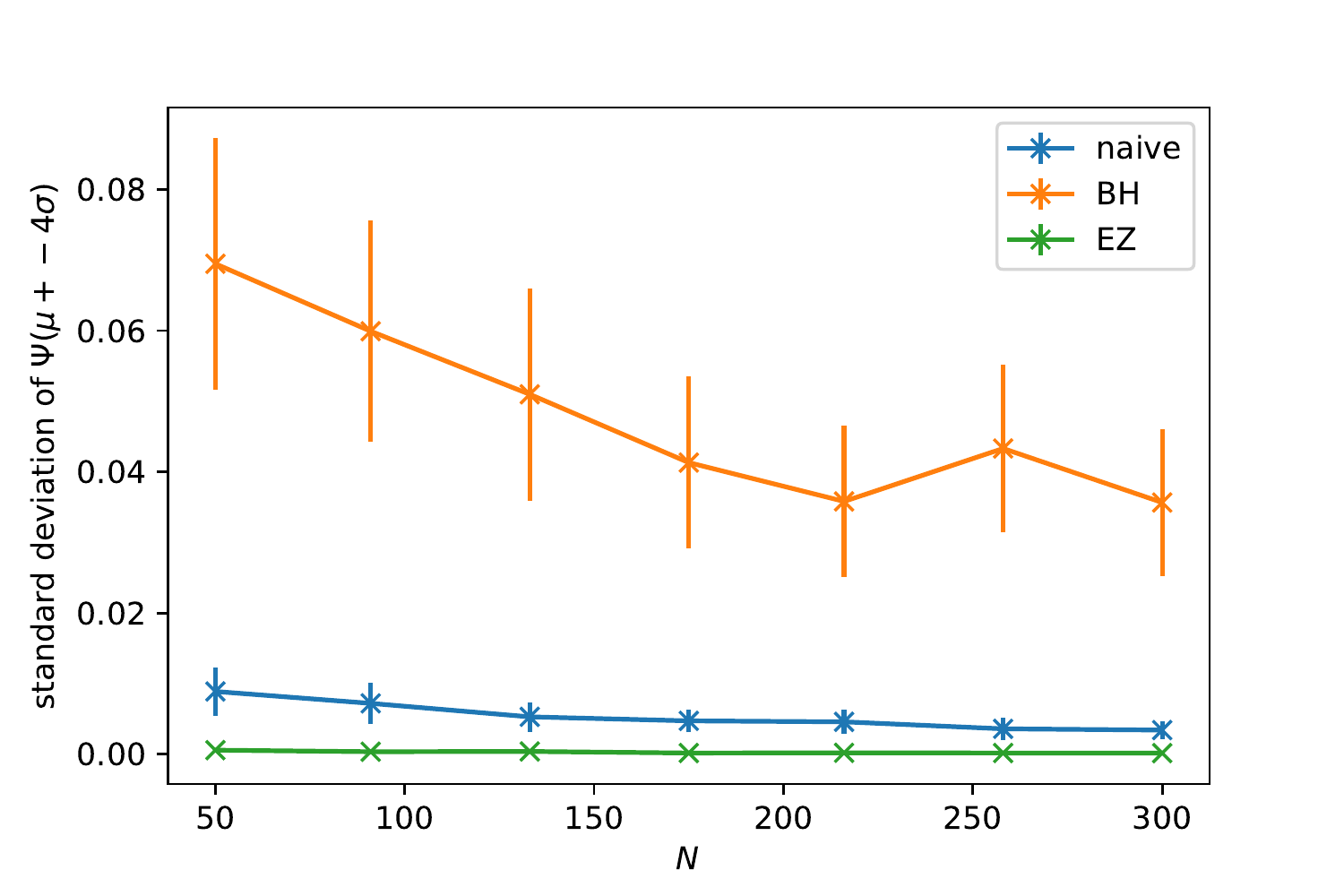}
        \caption{$y=\bar{\mu} + 4\bar{\sigma}$}
    \end{subfigure}
    \caption{Estimating the GP CDF $\Psi$ defined in (\ref{eq:psi_cdf}). Each plot shows results for na\"{i}ve, BC and EZ estimators using a range of sample sizes. We plot the standard deviation of the estimates over resampling of the DPP/i.i.d. points. The error bars show standard deviation over the 50 GP evaluation points. Each subfigure considers a different $y$ value in $\Psi(y)$.}
    \label{fig:gp_posterior_cdfs}
\end{figure}

\medskip
To really test the utility of our DPP MC integration approaches in this case, we must assess the accuracy of estimation of the CDF $\Psi$ defined in (\ref{eq:psi_cdf}).
To this end, for all $x$ values in the test set, we compute $\Psi(y)$ for each \begin{align}
    y \in \{\bar{\mu} - 4\bar{\sigma}, \bar{\mu} - 1.96\bar{\sigma}, \bar{\mu} - \bar{\sigma}, \bar{\mu}, \bar{\mu} + \bar{\sigma}, \bar{\mu} + 1.96\bar{\sigma}, \bar{\mu} + 4\bar{\sigma}\}.
\end{align}
Accurate estimation of such values is critical for accurate estimation of confidence intervals.
We repeat this estimation 30 times (resampling the DPP/i.i.d. sample points) leading to 30 values for each of the 7 $y$ values at each of the 50 test points.
To visualise, we then compute the sample standard deviation over the 30 re-samplings, leading to $7\times 50$ standard deviation values of which we take the sample mean over the 50 points.
Figure \ref{fig:gp_posterior_cdfs} shows the resulting standard deviation values for each of the 7 $y$ values and each of the estimators. 
The results show, perhaps surprisingly, EZ outperforming BH and na\"{i}ve in all cases.
This indicates that the GP posterior CDF has rapidly decaying coefficients in an Hermite polynomial expansion \cite{NEURIPS2019_1d54c76f}.

\section{Conclusions and future work}
We have considered multivariate Hermite polynomials for the construction of determinantal points processes with Gaussian base measure.
To enable their practical use, we have developed a novel multi-stage sampler for such Gauss-Hermite DPPs that is quite different from prior analogous samplers for other DPPs.
Our experimental timing and efficiency results demonstrate that the sampler is practically feasible and in-line with performance obtained by prior DPP samplers.
We have presented some experimental results using Gauss-Hermite DPPs in BH and EZ approaches to MC integration, showing strong performance in a variety of tasks.
In particular, we have presented some novel experimental results using Gauss-Hermite DPPs and BH/EZ estimation to marginalise the hyperparameters of a Gaussian process regression posterior.
The excellent results of the EZ estimator in this case suggest that the GP posterior, as functions of the raw hyperparameters, are well adapted to the Hermite basis.
We remark that the factorised form (\ref{eq:dpp_factorisation}) of the DPP posterior is clearly related to the posterior covariance of a Gaussian process with kernel $K_N$.
It would be interesting to consider if this similarity could be in part responsible for the excellent EZ performance, perhaps suggesting that EZ MC integration is particularly suited to GP posteriors.
Future work should also consider optimising the implementations to avoid repeated Hermite polynomials evaluations, in particular evaluations used in sampling from the top-level proposal density $f_N$ could be recycled for use in evaluating the acceptance ratio itself.
Finally, we constructed an empirically optimised rejection sampler for the univariate truncated Hermite mixtures $\rho_n$ being inspired by the convergence of such distributions to the Wigner semi-circle law.
This leaves open the question of obtaining analytically rejection bounds, and in particular bounds that are good enough for practical use in the rejection sampler.
We have not been able to derive such bounds, but the semi-circle convergence does suggest that they might be available. 

\bibliography{references} 

\begin{thebibliography}{10}

\bibitem{mehta2004random}
M.~L. Mehta, {\em Random matrices}.
\newblock Elsevier, 2004.

\bibitem{livan2018introduction}
G.~Livan, M.~Novaes, and P.~Vivo, {\em Introduction to random matrices: theory
  and practice}, vol.~26.
\newblock Springer, 2018.

\bibitem{hough2006determinantal}
J.~B. Hough, M.~Krishnapur, Y.~Peres, and B.~Vir{\'a}g, ``Determinantal
  processes and independence,'' {\em Probability surveys}, vol.~3,
  pp.~206--229, 2006.

\bibitem{NEURIPS2019_1d54c76f}
G.~Gautier, R.~Bardenet, and M.~Valko, ``On two ways to use determinantal point
  processes for monte carlo integration,'' in {\em Advances in Neural
  Information Processing Systems} (H.~Wallach, H.~Larochelle, A.~Beygelzimer,
  F.~d\textquotesingle Alch\'{e}-Buc, E.~Fox, and R.~Garnett, eds.), vol.~32,
  Curran Associates, Inc., 2019.

\bibitem{kulesza2012determinantal}
A.~Kulesza and B.~Taskar, ``Determinantal point processes for machine
  learning,'' {\em arXiv preprint arXiv:1207.6083}, 2012.

\bibitem{bardenet2020monte}
R.~Bardenet and A.~Hardy, ``Monte carlo with determinantal point processes,''
  {\em The Annals of Applied Probability}, vol.~30, no.~1, pp.~368--417, 2020.

\bibitem{ermakov1960polynomial}
S.~M. Ermakov and V.~Zolotukhin, ``Polynomial approximations and the
  monte-carlo method,'' {\em Theory of Probability \& Its Applications},
  vol.~5, no.~4, pp.~428--431, 1960.

\bibitem{krasikov2004new}
I.~Krasikov, ``New bounds on the hermite polynomials,'' {\em arXiv preprint
  math/0401310}, 2004.

\bibitem{anderson2010introduction}
G.~W. Anderson, A.~Guionnet, and O.~Zeitouni, {\em An introduction to random
  matrices}.
\newblock No.~118, Cambridge university press, 2010.

\bibitem{gardner2018gpytorch}
J.~R. Gardner, G.~Pleiss, D.~Bindel, K.~Q. Weinberger, and A.~G. Wilson,
  ``Gpytorch: Blackbox matrix-matrix gaussian process inference with gpu
  acceleration,'' in {\em Advances in Neural Information Processing Systems},
  2018.

\bibitem{lalchand2020approximate}
V.~Lalchand and C.~E. Rasmussen, ``Approximate inference for fully bayesian
  gaussian process regression,'' in {\em Symposium on Advances in Approximate
  Bayesian Inference}, pp.~1--12, PMLR, 2020.

\bibitem{dunkl2014orthogonal}
C.~F. Dunkl and Y.~Xu, {\em Orthogonal polynomials of several variables}.
\newblock No.~155, Cambridge University Press, 2014.

\bibitem{lohofer1991inequalities}
G.~Loh{\"o}fer, ``Inequalities for legendre functions and gegenbauer
  functions,'' {\em Journal of Approximation Theory}, vol.~64, no.~2,
  pp.~226--234, 1991.

\end{thebibliography}
% \printbibliography

\appendix

\section{Gaussian orthonormal functions from spherical orthogonal functions}\label{app:spherical}
This section describes another novel construction of Guassian orthogonal function on $\R^N$ using spherical orthogonal functions.
It is included for interest and to illustrate the difficulty of sampling from distributions of the form $\phi_{\vec{i}(\x)}d\x$ in the Gaussian case, as opposed to the Jacobi case of \cite{NEURIPS2019_1d54c76f} which is well-behaved.

\medskip
Suppose that $\{\psi_{\n}\}_{\n}$ are a set of orthonormal functions defined on the $d$-sphere (that is, the $d-1$ dimensional surface that is embedded in $\R^d$).
The multi-indices $\n$ are $d$-tuples of non-negative integers.
We have the defining orthonormality property:
\begin{equation*}
    \int_{S^d} dS(\vec{e}) ~ \psi_{\n}(\vec{e}) \psi_{\m}(\vec{e}) = \delta_{\n\m},
\end{equation*}
where $dS$ is the surface measure on the sphere. Let $\sigma_d$ be the surface area of the $d$-sphere, then
\begin{align*}
    \int_{\R^d} d\vec{x} ~ \mathcal{N}(\vec{x}; 0, I_d) \psi_{\n}\left(\frac{\vec{x}}{\|\vec{x}\|_2}\right)\psi_{\m}\left(\frac{\vec{x}}{\|\vec{x}\|_2}\right) = \int_0^{\infty} dr ~ r^{d-1} \frac{e^{-\frac{r^2}{2}}}{(2\pi)^{d/2}} \int_{S^d} dS(\vec{e}) \psi_{\n}(\vec{e})\psi_{\m}(\vec{e}) = \sigma_{d}^{-1}\delta_{\n\m}.
\end{align*}
Define $\phi_{\n}(\x) = \sqrt{\sigma_d} \psi_{\n}\left(\frac{\vec{x}}{\|\vec{x}\|_2}\right)$. Then we have shown $\{\phi_{\n}\}_{\n}$ are an orthonormal set of functions with respect to the standard Gaussian base measure on $\R^d$.

\medskip 

A construction of orthogonal polynomials on the $d$-sphere is provided by \cite{dunkl2014orthogonal}.
Here we simply repeat the construction, however the literature contains various conventions for classical orthogonal polynomials, so we will take particular care ensure the correct normalisation.

The orthogonal polynomials are constructed using Chebyshev polynomials and Gegenbauer polynomials. 
One can verify that the Gegenbauer polynomials defined in \cite{dunkl2014orthogonal} match those provided by the \texttt{scipy} implementation.
Let $C_{n^{(\lambda)}}$ denote the Gegenbauer polynomials. Let also $T_s, U_s$ denote the Chebyshev polynomials of the first and second kinds respectively.
Then define $g_{s, 0}(\vec{x}) = \rho^s T_s(x_{d-1}\rho^{-1})$ and $g_{s, 1}(\vec{x}) =  x_d\rho^{s} U_s(x_{d-1}\rho^{-1})$ where $\rho^2 = x_{d-1}^2 + x_d^2$.
We can then define the orthogonal polynomials from \cite{dunkl2014orthogonal}:
\begin{align}
    \chi_{\n}(\vec{x}) = g_{n_{d-1}, n_d}(\vec{x}) \prod_{j=1}^{d-2}\left( r_j(\vec{x})^{2n_j} C_{n_j}^{(\lambda_j)}\left(x_jr_j(\vec{x})^{-1}\right)\right)
\end{align}
where $r_j(\vec{x})^2 = x_j^2 + \ldots + x_d^2$ and $\lambda_j = \frac{d-j-1}{2} + \sum_{i=j+1}^{d}n_i$.
\cite{dunkl2014orthogonal} also provides the normalisation, so we define 
\begin{align}
    \psi_{\n}(\vec{x}) = \frac{1}{\sqrt{\sigma_da_{\n}}} \left[\prod_{j=1}^{d-2}\sqrt{\frac{n_j! \left(\frac{d-j+1}{2}\right)_{\beta_j} (n_j + \lambda_j)}{\lambda_j (2\lambda_j)_{n_j}\left(\frac{d-j}{2}\right)_{\beta_j}}} \right]\chi_{\n}(\vec{x})
\end{align}
where $a_{\n} = \frac{1}{2}$ is $n_{d-1} + n_d > 0$, else $1$, and $\beta_j=\sum_{i=j+1}^d n_i$. So the orthonormal functions are
\begin{align}
    \phi_{\n}(\vec{x}) = \frac{1}{\sqrt{a_{\n}}} \left[\prod_{j=1}^{d-2}\sqrt{\frac{n_j! \left(\frac{d-j+1}{2}\right)_{\beta_j} (n_j + \lambda_j)}{\lambda_j (2\lambda_j)_{n_j}\left(\frac{d-j}{2}\right)_{\beta_j}}} \right]\chi_{\n}\left(\frac{\vec{x}}{\|\vec{x}\|_2}\right).
\end{align}

\cite{lohofer1991inequalities} provides a bound for the Gegenbauer polynomials. The normalisation conventions are not specified, but one can easily verify numerically that they match those used above. We have therefore:
\begin{align}
    |C_{n}^{(\lambda)}(x)| \leq c_{2n, 2\lambda} x^2 + c_{n, \lambda} (1-x^2), ~~ c_{n, \lambda} = \frac{\Gamma(n/2 + \lambda)}{\Gamma(\lambda)\Gamma(n/2 + 1)}.
\end{align}

The Chebyshev polynomials are well-known to be uniformly bounded by 1, so we obtain 
\begin{align}
    |\phi_{\n}(\vec{x})|^2 \leq \frac{1}{a_{\n}} \prod_{j=1}^{d-2}\frac{n_j! \left(\frac{d-j+1}{2}\right)_{\beta_j} (n_j + \lambda_j)\Gamma(n_j + 2\lambda_j)^2}{\lambda_j (2\lambda_j)_{n_j}\left(\frac{d-j}{2}\right)_{\beta_j}\Gamma(2\lambda_j)^2\Gamma(n_j + 1)^2} &= \frac{1}{a_{\n}} \prod_{j=1}^{d-2}\frac{\left(\frac{d-j+1}{2}\right)_{\beta_j} (n_j + \lambda_j)(2\lambda_j)_{n_j}}{\lambda_j \left(\frac{d-j}{2}\right)_{\beta_j}n_j \Gamma(n_j)}\notag\\
     &= \frac{1}{a_{\n}} \prod_{j=1}^{d-2} \frac{(n_j + \lambda_j) B\left( \frac{d-j}{2} + \beta_j, \frac{1}{2}\right)B(2\lambda_j, n_j)}{\lambda_jn_j B\left( \frac{d-j}{2}, \frac{1}{2}\right)}
\end{align}
where the final equality comes from simple Gamma function manipulations.

\medskip
Following \cite{NEURIPS2019_1d54c76f} we must construct a rejection sampler for $\phi_{\n}(\cdot)^2 \mathcal{N}(\cdot; 0, I_d)$
The obvious proposal distribution is $\omega(\cdot) = \mathcal{N}(\cdot; 0, I_d)$ and then the previous section's results give
\begin{align}
    \frac{\phi_{\n}(\vec{x})^2 \mathcal{N}(\vec{x}; 0, I_d)}{\omega(\vec{x})} \leq \frac{1}{a_{\n}} \prod_{j=1}^{d-2} \frac{(n_j + \lambda_j) B\left( \frac{d-j}{2} + \beta_j, \frac{1}{2}\right)}{\lambda_j B\left( \frac{d-j}{2}, \frac{1}{2}\right)} \equiv M_{\n}.
\end{align}
Overall, samples $\vec{x}$ are proposed from $\omega$ and are rejected if $\phi_{\n}(\vec{x})^2 < M_{\n} u$ for $u\sim U[0,1]$.

\medskip 
This approach is not particularly viable however, as the constants $M_{\vec{n}}$ are shown in practice to scale extremely badly with $\vec{n}$ and $d$.
Moreover, the discrepancy between the minima of $\phi_{\n}(\vec{x})^2 \mathcal{N}(\vec{x}; 0, I_d)$ and its upper bound is extreme.
These two facts combine to give extremely poor acceptance rates for even modest values of $N$ and $d$.

\end{document}